\documentclass[journal]{IEEEtran}
\usepackage{amsmath,amsfonts}
\usepackage{algorithmic}
\usepackage{algorithm}
\usepackage{array}
\usepackage[caption=false,font=normalsize,labelfont=sf,textfont=sf]{subfig}
\usepackage{textcomp}
\usepackage{stfloats}
\usepackage{url}
\usepackage{verbatim}
\usepackage{graphicx}
\usepackage{cite}
\usepackage{multirow}
\usepackage{graphicx}
\usepackage{xcolor}
\begin{document}

\title{A Minimal Solver for Relative Pose Estimation with Unknown Focal Length from Two Affine Correspondences}

\author{Zhenbao Yu, Shirong  Ye, Ronghe Jin, Shunkun Liang, Zibin Liu, Huiyun Zhang, Banglei Guan
\thanks{Manuscript received: July 15, 2025; Revised: Septembe 24; Accepted November 26, 2025.}
\thanks{This paper was recommended for publication by Editor Pascal Vasseur upon evaluation of the Associate Editor and Reviewers’ comments.
{Corresponding author: Shirong  Ye.}} 
\thanks{Zhenbao Yu, Shunkun Liang, Zibin Liu, Banglei Guan are with the School of Aerospace Science and Engineering, National University of Defense Technology, and Hunan Provincial Key Laboratory of Image Measurement and Vision Navigation, Changsha 410000, China.
{\tt\footnotesize zhenbaoyu@whu.edu.cn, liangshunkun@nudt.edu.cn, liuzibin19@nudt.edu.cn, banglei.guan@hotmail.com}}

\thanks{Shirong Ye is with the Global Navigation Satellite System Research Center, and the School of Geodesy and Geomatics, Wuhan University, Wuhan 430000, China.
{\tt\footnotesize srye@whu.edu.cn}}
\thanks{Ronghe Jin is with the School of Remote Sensing Information Engineering, Wuhan University, Wuhan 430000, China.
{\tt\footnotesize huanhexiao@whu.edu.cn}}
\thanks{Huiyun Zhang is with the School of Software, Henan University, Kaifeng 475004, China.
{\tt\footnotesize zhzhy@henu.edu.cn}}

\thanks{Digital Object Identifier (DOI): see top of this page.}
}

\markboth{****************, ***************}
{*****************} 
\maketitle
\IEEEpubid{0000--0000/00\$00.00~\copyright~2025 IEEE}
\begin{abstract}
In this paper, we aim to estimate the relative pose and focal length between two views with known intrinsic parameters except for an unknown focal length from two affine correspondences (ACs). Cameras are commonly used in combination with inertial measurement units (IMUs) in applications such as self-driving cars, smartphones, and unmanned aerial vehicles. The vertical direction of camera views can be obtained by IMU measurements. The relative pose between two cameras is reduced from 5DOF to 3DOF. We propose a new solver to estimate the 3DOF relative pose and focal length. First, we establish constraint equations from two affine correspondences when the vertical direction is known. Then, based on the properties of the equation system with nontrivial solutions, four equations can be derived. These four equations only involve two parameters: the focal length and the relative rotation angle. Finally, the polynomial eigenvalue method is utilized to solve the problem of focal length and relative rotation angle. The proposed solver is evaluated using synthetic and real-world datasets. The results show that our solver performs better than the existing state-of-the-art solvers. 
\end{abstract}


\section{Introduction}
\IEEEPARstart{T}{he} task of estimating the relative pose of camera motion from two views is a foundational task in computer vision. It plays an important role in many popular areas, such as simultaneous localization and mapping (SLAM)~\cite{SLAM2_RAL,SLAM3_RAL}, structure-from-motion (SFM)~\cite{SMF1,SFM2,SFM_TMM}, and visual odometry (VO)~\cite{VO,VO_tim,vO2_tim}. Improving the accuracy, robustness, and numerical stability of relative pose estimation remains an active research topic.

The essential and fundamental matrix is usually used to represent the relative poses of calibrated and uncalibrated cameras, respectively~\cite{book}. Estimation of the essential matrix for a calibrated camera requires 5 point correspondences~\cite{5PT1}, in contrast to the 7 or 8 needed for the fundamental matrix with an uncalibrated camera~\cite{book}. The intrinsic parameters of the camera include focal length, aspect ratio, principal point, and non-perspective distortion parameters. It is a very common assumption that the camera has known intrinsic parameters except for an unknown common focal length (semi-calibrated). This assumption is reasonable since (1) the principal point of CCD or CMOS cameras is located in the center of the photo, and the aspect ratio is usually 1:1; (2) the distortion can be ignored if the field of view is narrow~\cite{AC2}. Besides, the focal length usually remains consistent during two consecutive image shots~\cite{stewenius_21,lihongdong}.

The relative pose and focal length are obtained by using 6 point correspondences if the focal lengths of the two cameras are equal~\cite{stewenius,lihongdong,Polynomial}. These methods, which use the minimal number of samples to estimate the relative pose, are known as \textit{minimal solver}. To handle outlier correspondences from cluttered environments, robust estimators like Random Sample Consensus (RANSAC)~\cite{Random} are typically employed with a \textit{minimal solver}. Within the RANSAC framework, the number of iterations and the minimum number of samples required for \textit{minimal solver} are negatively correlated~\cite{saurer2016homography}. Therefore, reducing this sample size is vital for enhancing both efficiency and accuracy.

Aiming to reduce the minimal sample size, a series of studies have proposed affine correspondences for relative pose estimation~\cite{barath_homography,barath_planar,AC2,barath2018Essential,guan1}. An affine correspondence (AC), comprising a point correspondence and a local affine transformation, provides three independent equations in total: one from the point pair and two from the affine component~\cite{AC2}. In practice, the camera and IMU often form a fixed sensor assembly, a configuration prevalent in devices including unmanned aerial vehicles (UAVs), tablets, and smartphones~\cite{IMU1,IMU2,IMU3,IMU4}. With the roll and pitch angles supplied by an IMU, the relative rotation's degrees of freedom are reduced from 3 to 1. For cameras with a common reference direction, a minimal solution using four point correspondences to estimate the relative pose and focal length is presented in~\cite{ding_f}. The external parameter calibration of cameras and IMUs can refer to existing literature~\cite{Yang2017Monocular,2013Unified}.

\IEEEpubidadjcol
We are the first to establish a monocular relative pose and focal length solver requiring only two affine correspondences with an IMU vertical direction. Our approach advances prior work in three key aspects: incorporating an IMU prior to achieve 3-DOF estimation with a stable polynomial eigenvalue solver beyond~\cite{AC2}; handling arbitrary camera motions unlike the planar motion restriction in~\cite{guan1}; and reducing the required sample size from 4 point correspondences in~\cite{ding_f} to merely 2 ACs under identical IMU constraints. The main contributions of this work are summarized below.

$ \bullet \, $ Constraint equations are established from two affine correspondences with an unknown focal length when the direction of the vertical direction is known. Based on the properties of the equation system with nontrivial solutions, four equations containing only the parameters of the focal length and the relative rotation angle can be derived.

$ \bullet \, $ The maximum degree of the focal length and the relative rotation angle are 5 and 6. The polynomial eigenvalue method is utilized to solve the problem of focal length and relative rotation angle. The translation vector is extracted from the corresponding eigenvalues.

$ \bullet \, $ The superiority of our solver in terms of numerical stability and accuracy compared with the state-of-the-art solvers is verified using synthetic data and real data.

The rest of this paper is organized as follows. We review the related work in Section 2. Geometric constraints are introduced in Section 3. In Section 4, a solution for estimating the relative pose (3DOF) and focal length of cameras using two affine correspondences is proposed. Furthermore, we test the performance of the proposed method and the comparison method in Section 5, and the conclusions in Section 6.

\section{Related work}
The solution of calculating the relative pose and focal length of the camera from 6 point correspondences is proposed in~\cite{stewenius}. The practical application of this solution, which uses the Gröbner basis method, is challenging because it requires a solid foundation in algebraic geometry and computational commutative algebra for accurate results. Therefore, a new solution based on the hidden-variable technique is proposed on the basis of 6 point correspondences~\cite{lihongdong}. To overcome the failure of the 6-point method under critical motions (e.g., pure rotation or translation)~\cite{stewenius_10}, it is extended by a 4+1 plane-parallax solver in~\cite{problem_6pt}. For the common case of a semi-calibrated camera equipped with an IMU, the relative pose and focal length can be estimated with only four point correspondences~\cite{ding_f,ding_H}.

In recent years, there have been a number of methods proposed using affine correspondence. An explicit relation between local affine approximations by from matching of affine invariant regions and the epipolar geometry is proposed~\cite{bentolila2014conic}. Estimating the homography and essential matrix using affine correspondences is proposed in~\cite{SFM2}. Linear constraints between the local affine transformation and the essential matrix are proposed in~\cite{barath2018Essential}. The estimation of the relative pose using 2ACs for any central camera model~\cite{eichhardt2018affine}. For planar motion and unknown focal length, a new solver is provided to estimate the relative pose and focal length of the camera based on Gröbner basis from 1AC~\cite{guan1}. The relative pose and focal length can be estimated from two semi-calibrated cameras using 2ACs, which provide six independent constraints. These constraints are then solved via the hidden variable technique~\cite{AC2}.

As most minimal solvers generate nontrivial polynomial systems, their efficient and accurate solution is critical for relative pose estimation, primarily achieved using Gröbner basis and polynomial ideal theory.The Gröbner basis is used in minimal solution problems such as the well-known 5-pt~\cite{larsson2018beyond} for calibrated cameras and 6-pt for semi-calibrated~\cite{stewenius}. The lack of portability in Gröbner basis methods, which necessitates designing a new solver for each problem to ensure efficiency and robustness, is addressed by an automatic solver generator~\cite{kukelova2008Gröbner}. However, additional expertise is usually required when dealing with more difficult problems. Solving polynomial equations by polynomial eigenvalue is proposed in the field of computer vision~\cite{eigenvalue}. The polynomial eigenvalue method is easier to implement, providing direct solvers for the 5-pt~\cite{Polynomial}, 6-pt~\cite{Polynomial}, and 9-pt problems~\cite{fitzgibbon2001simultaneous}.  Polynomial eigenvalue and Gröbner basis solvers for 3D reconstruction with known focal length are developed in~\cite{kukelova20093d}.

Deep learning has enabled new methodologies for estimating relative pose. Learning-based techniques for estimating frame-to-frame motion fields are proposed in~\cite{Dosovitskiy2015FlowNetLO,mayer2016large}. SfM-Net decomposes per-frame pixel motion based on scene and object depth, camera motion, and 3D object transformations~\cite{vijayanarasimhan1704sfm}. The work in~\cite{wang2021deep} presents an SfM literature review and a proposed deep learning-based two-view framework. Recent advances include a method integrating traditional geometry with deep learning~\cite{zhuang2021fusing}, along with the end-to-end NFlowNet for direct camera pose estimation~\cite{parameshwara2022diffposenet}. Deep neural networks enable accurate estimation of each affine component's parameters, yielding excellent performance in image matching and relative pose estimation~\cite{sun2025learning}. Despite the promise of deep learning, it remains limited by generalization issues and substantial resource requirements. In comparison, the geometric approach explored here offers greater interpretability. In this paper, we mainly focus on estimating the relative pose and focal length of the camera by affine correspondence.

\section{Geometric Constraint}
Suppose there is a 3D point ${{\bf{X}}}$ observed in two views. Two corresponding image points ${{\bf{x}}_i}$ and ${{\bf{x}}_j}$ are obtained for a 3D point in ${i}$ and ${j}$ views, respectively. The relationship between fundamental matrix ${\bf{F}}$ and image points can be expressed as
\begin{equation}
	\begin{aligned}
{{{\bf{x}}_j}^{\rm{T}}}{\bf{F}}{{\bf{x}}_i} = 0
 	\end{aligned}.
	\label{eq1}
\end{equation}

The reciprocal of the focal length is ${f}$. For semi-calibrated cameras~\cite{ding_f,AC2,lihongdong}, the camera intrinsic matrix ${{\bf{K}}}$ can be expressed as
\begin{equation}
	\begin{aligned}
{{\bf{K}^{ - 1}}} = diag({1},{1},f)
 	\end{aligned}.
	\label{eq2}
\end{equation}

The relationship between the fundamental matrix ${\bf{F}}$ and the essential matrix ${\bf{E}}$ is given by
\begin{equation}
	\begin{aligned}
{\bf{F}} = {\bf{K}}^{\rm{- T} }{\bf{EK}}^{ - 1}
 	\end{aligned}.
	\label{eq3}
\end{equation}

Substituting~\eqref{eq3} and~\eqref{eq2} into~\eqref{eq1} yields
\begin{equation}
	\begin{aligned}
{({\bf{K}}^{ - 1}{{\bf{x}}_j})^{\rm{T}}}{\bf{EK}}^{ - 1}{{\bf{x}}_i} = 0
 	\end{aligned},
	\label{eq4}
\end{equation}
\begin{equation}
	\begin{aligned}
{\bf{E}} = {\left[ {\bf{t}} \right]_ \times }{\bf{R}}
 	\end{aligned},
	\label{eq5}
\end{equation}
where ${\bf{R}}$, ${\bf{t}}$ represent the rotation matrix and translation vector from ${i}$ to ${j}$ views.

\subsection{Relationship of Fundamental Matrix and Local Affinities}

An affine correspondence consists of a local affine transformation ($\bf A$) and a point correspondence (${{\bf{x}}_i},{{\bf{x}}_j}$). The relationship of the fundamental matrix and local affine transformation can be expressed as~\cite{AC2}:
\begin{equation}
	\begin{aligned}
{({{{\bf{\hat A}}}^{ - {\rm{T}}}}{{\bf{F}}^{\rm{T}}}{{\bf{x}}_j})_{(1:2)}} =  - {({{\bf{F}}}{{\bf{x}}_i})_{(1:2)}}
 	\end{aligned}.
	\label{eq10}
\end{equation}

Substituting~\eqref{eq3} into~\eqref{eq10}, we get
\begin{equation}
	\begin{aligned}
{({{\bf{\hat A}}^{ - {\rm{T}}}}{\bf{K}}^{ - {\rm{T}}}{{\bf{E}}^{\rm{T}}}{\bf{K}}^{ - 1}{{\bf{x}}_j})_{(1:2)}} =  - {({\bf{K}}^{ - {\rm{T}}}{{\bf{E}}}{{\bf{K}}^{ - 1}}{{\bf{x}}_i})_{(1:2)}}
 	\end{aligned},
	\label{eq11}
\end{equation}
where $\bf{\hat A}$ is a $3 \times 3$ matrix as follows:
\begin{equation}
	\begin{aligned}
{\bf{\hat A}} = \left[ {\begin{array}{*{20}{c}}
{\bf{A}}&{\bf{0}}\\
{\bf{0}}&1
\end{array}} \right]
 	\end{aligned}.
	\label{eq12}
\end{equation}

The local affine transformation $\bf{A}$ is a $2 \times 2$ matrix which is to the vicinity of point ${{\bf{x}}_i}$ to that of ${{\bf{x}}_j}$.  The affine matrix A can be obtained by using the ASIFT algorithm~\cite{ASIFT}.
\begin{equation}
	\begin{aligned}
{\bf{A}} = \left[ {\begin{array}{*{20}{c}}
{{a_{11}}}&{{a_{12}}}\\
{{a_{21}}}&{{a_{22}}}
\end{array}} \right]
 	\end{aligned}.
	\label{eq13}
\end{equation}

\section{Relative Pose Estimation With Known Vertical Direction and Unknown Focal Length}
\subsection{Problem formulation}
The IMU provides the roll (${\theta _{\rm{z}}}$) and pitch (${\theta _{\rm{x}}}$) angles of the camera. Using the pitch and roll angles, we align the camera views with the vertical direction. We assume that the positive direction of the Y-axis of the camera coordinate system is vertical direction, and the X-Z plane is orthogonal to the Y-axis. The rotation matrix ${{\bf{R}}_{{\rm{imu}}}}$ is used to align the camera coordinate system and the vertical direction. ${{\bf{R}}_{{\rm{imu}}}}$ can be written as

\setlength{\arraycolsep}{1.2pt}
\begin{equation}
	\begin{aligned}
    \small{
{{\bf{R}}_{{\rm{imu}}}} = \underbrace {\left[ {\begin{array}{*{20}{c}}
1&0&0\\
0&{\cos ({\theta _x})}&{\sin ({\theta _x})}\\
0&{ - \sin ({\theta _x})}&{\cos ({\theta _x})}
\end{array}} \right]}_{{{\bf{R}}_x}}\underbrace {\left[ {\begin{array}{*{20}{c}}
{\cos ({\theta _z})}&{\sin ({\theta _z})}&0\\
{ - \sin ({\theta _z})}&{\cos ({\theta _z})}&0\\
0&0&1
\end{array}} \right]}_{{{\bf{R}}_z}}}
 	\end{aligned}.
	\label{eq14}
\end{equation}

Define ${{\bf{R}}_i}$ and ${{\bf{R}}_j}$ as the rotation matrix of views ${i}$ and ${j}$ relative to the original coordinate system. We can get

\begin{equation}
	\begin{aligned}   
       \left\{ \begin{array}{l}
     	 {\bf{R}} = {{\bf{R}}_j^{T}}{{\bf{R}}_i} = {({\bf{R}}_{{\rm{imu}}}^j)^T}{{\bf{R}}_y}{\bf{R}}_{{\rm{imu}}}^i,\\
     	{\bf{t}} = {\left( {{\bf{R}}_x^j{\bf{R}}_z^j} \right)^T}{\bf{\tilde t}} = {\left( {{\bf{R}}_{{\rm{imu}}}^j} \right)^T}{\bf{\tilde t}},
      \end{array} \right.
	\end{aligned}
	\label{eq15}
\end{equation}
where ${{\bf{R}}_y} = {({\bf{R}}_y^j)^T}{\bf{R}}_y^i$. ${\bf{R}}_{{\rm{imu}}}^i$ and ${\bf{R}}_{{\rm{imu}}}^j$ as the rotation matrices provided by the IMU in view ${i}$ and ${j}$. ${{\bf{R}}_y}$ and ${\bf{\tilde t}}$ are the rotation matrix and translation vector between the aligned views ${i}$ and ${j}$.
\begin{equation}
	\begin{aligned}
    {{\bf{R}}_y} = \left[ {\begin{array}{*{20}{c}}
    {{\rm{cos}}\left( \theta  \right)}&0&{{\rm{sin}}\left( \theta  \right)}\\
     0&1&0\\
     {{\rm{ - sin}}\left( \theta  \right)}&0&{{\rm{cos}}\left( \theta  \right)}
      \end{array}} \right],{\bf{\tilde t}} = \left[ {\begin{array}{*{20}{c}}
      {{{\tilde t}_x}}\\
      {{{\tilde t}_y}}\\
      {{{\tilde t}_z}}
        \end{array}} \right],
 	\end{aligned}
	\label{eq16}
\end{equation}
where ${\theta}$ is the rotation angle around the Y-axis, which can be rewritten using Cayley parameterization as:
\begin{equation}
	\begin{aligned}
       {{\bf{R}}_y} = \frac{1}{{1 + {s^2}}}\left[ {\begin{array}{*{20}{c}}
	     	{1 - {s^2}}&0&{2s}\\
	        	0&{1 + {s^2}}&0\\
	    	{ - 2s}&0&{1 - {s^2}}
      \end{array}} \right],
	\end{aligned}
	\label{eq17}
\end{equation}
where $ s= \tan \frac{\theta }{2}$.

By substituting~\eqref{eq15} into~\eqref{eq5}, we can get
\begin{equation}
	\begin{aligned}
{\bf{E}} = {\left[ {{{\left( {{\bf{R}}_{{\rm{imu}}}^j} \right)}^T}{\bf{\tilde t}}} \right]_ \times }{({\bf{R}}_{{\rm{imu}}}^j)^T}{{\bf{R}}_{\bf{y}}}{\bf{R}}_{{\rm{imu}}}^i = {({\bf{R}}_{{\rm{imu}}}^j)^T}{\bf{\tilde ER}}_{{\rm{imu}}}^i
	\end{aligned},
	\label{eq18}
\end{equation}
where  ${\bf{\tilde E}} = {\left[ {{\bf{\tilde t}}} \right]_ \times }{{\bf{R}}_y}$.

By substituting Eq.~\eqref{eq18} into Eq.~\eqref{eq4}, we get
\begin{equation}
	\begin{aligned}
{({\bf{R}}_{{\rm{imu}}}^j{{\bf{K}}^{ - 1}}{{\bf{x}}_j})^{\rm{T}}}{\bf{\tilde ER}}_{{\rm{imu}}}^i{{\bf{K}}^{ - 1}}{{\bf{x}}_i} = 0
	\end{aligned}.
	\label{eq19}
\end{equation}

By substituting Eq.~\eqref{eq18} into Eq.~\eqref{eq11}, we get
\begin{equation}
	\begin{aligned}
{({{{\bf{\hat A}}}^{ - {\rm{T}}}}{{\bf{K}}^{ - {\rm{T}}}}{({\bf{R}}_{{\rm{imu}}}^i)^{\rm{T}}}{{{\bf{\tilde E}}}^{\rm{T}}}{\bf{R}}_{{\rm{imu}}}^j{{\bf{K}}^{ - 1}}{{\bf{x}}_j})_{(1:2)}} = \\ - {({{\bf{K}}^{ - {\rm{T}}}}{({\bf{R}}_{{\rm{imu}}}^j)^{\rm{T}}}{\bf{\tilde ER}}_{{\rm{imu}}}^i{{\bf{K}}^{ - 1}}{{\bf{x}}_i})_{(1:2)}}
	\end{aligned}.
	\label{eq20}
\end{equation}

 According to~\eqref{eq19} and~\eqref{eq20}, it is easy to find that an affine correspondence can provide three independent equations. We choose three equations from one affine correspondence and one equation from another point correspondence. We can get the following equation
\begin{equation}
	\begin{aligned}
\frac{1}{{1 + {s^2}}}\underbrace {\left[ {\begin{array}{*{20}{c}}
{{{\bf{M}}_{11}}}&{{{\bf{M}}_{12}}}&{{{\bf{M}}_{13}}}\\
{{{\bf{M}}_{21}}}&{{{\bf{M}}_{22}}}&{{{\bf{M}}_{23}}}\\
{{{\bf{M}}_{31}}}&{{{\bf{M}}_{32}}}&{{{\bf{M}}_{33}}}\\
{{{\bf{M}}_{41}}}&{{{\bf{M}}_{42}}}&{{{\bf{M}}_{43}}}
\end{array}} \right]}_{\bf{M}}\left[ \begin{array}{l}
{{{\bf{\tilde t}}}_x}\\
{{{\bf{\tilde t}}}_y}\\
{{{\bf{\tilde t}}}_z}
\end{array} \right] = {\bf{0}}
	\end{aligned},
	\label{eq21}
\end{equation}
where ${\bf{M}}$ is $4 \times 3$ matrix.

Since the equation system~\eqref{eq21} admits nontrivial solutions, the determinant of any three rows in ${\bf{M}}$ is zero. We can get ${\rm{C}}_4^3 = 4$ equations
\begin{equation}
	\begin{aligned}
\left\{ \begin{array}{l}
{{\rm{g}}_{\rm{1}}}(s,f){\rm{ = }}\det ({{\bf{M}}_{123}}) = 0\\
{{\rm{g}}_{\rm{2}}}(s,f){\rm{ = }}\det ({{\bf{M}}_{124}}) = 0\\
{{\rm{g}}_{\rm{3}}}(s,f){\rm{ = }}\det ({{\bf{M}}_{134}}) = 0\\
{{\rm{g}}_{\rm{4}}}(s,f){\rm{ = }}\det ({{\bf{M}}_{234}}) = 0
\end{array} \right.
	\end{aligned},
	\label{eq22}
\end{equation}
where ${{\bf{M}}_{ijk}}$ represents rows ${i}$, ${j}$, ${k}$ of matrix ${\bf{M}}$. The highest degree term is ${{s^6}{f^4}}$ in ${{\rm{g}}_{\rm{1}}}(s,f)$ and ${{\rm{g}}_{\rm{2}}}(s,f)$. The highest degree term is ${{s^6}{f^5}}$ in ${{\rm{g}}_{\rm{3}}}(s,f)$ and ${{\rm{g}}_{\rm{4}}}(s,f)$.
\subsection{Polynomial eigenvalue solver}
In this section, the polynomial eigenvalue solver is introduced for solving ${\bf{R}}$ , ${\bf{t}}$ and ${f}$. We rewrite~\eqref{eq22} to get
\begin{equation}
\left\{ {\begin{array}{*{20}{c}}
{\bf{C}}1*{\bf{X}}_1 = 0\\
{\bf{C}}2*{\bf{X}}_2 = 0\\
{\bf{C}}3*{\bf{X}}_3 = 0\\
{\bf{C}}4*{\bf{X}}_4 = 0
\end{array}} \right.
\label{eq23},
\end{equation}
where ${\bf{X}}_1={\bf{X}}_2$ and ${\bf{X}}_3={\bf{X}}_4$, which can be expressed as
\begin{equation}
      \begin{aligned}
{\bf{X}}_1 =& (1,f,{f^2},{f^3},{f^4},s,sf,s{f^2},s{f^3},s{f^4},{s^2},{s^2}f,{s^2}{f^2},\\&{s^2}{f^3},{s^2}{f^4},{s^3},{s^3}f,{s^3}{f^2},{s^3}{f^3},{s^3}{f^4},{s^4},{s^4}f,{s^4}{f^2},\\&{s^4}{f^3},{s^4}{f^4},{s^5},{s^5}f,{s^5}{f^2},{s^5}{f^3},{s^5}{f^4},\\&{s^6},{s^6}f,{s^6}{f^2},{s^6}{f^3},{s^6}{f^4})^T
	\end{aligned}.
\label{eq24}
\end{equation}
\begin{equation}
	\begin{aligned}
{\bf{X}}_3 = &(1,f,{f^2},{f^3},{f^4},{f^5},s,sf,s{f^2},s{f^3},s{f^4},s{f^5},{s^2},{s^2}f,\\&{s^2}{f^2},{s^2}{f^3},{s^2}{f^4},{s^2}{f^5},{s^3},{s^3}f,{s^3}{f^2},{s^3}{f^3},{s^3}{f^4},{s^3}{f^5}\\&{s^4},{s^4}f,{s^4}{f^2},{s^4}{f^3},{s^4}{f^4},{s^4}{f^5},{s^5},{s^5}f,{s^5}{f^2},{s^5}{f^3},{s^5}{f^4},\\&{s^5}{f^5},{s^6},{s^6}f,{s^6}{f^2},{s^6}{f^3},{s^6}{f^4},{s^6}{f^5})^T
	\end{aligned}.
	\label{eq25}
\end{equation}

~\eqref{eq23} can be rewritten as 
\begin{equation}
	\begin{aligned}
{{\bf{B}}(s)}{{\bf{J}}} = 0
	\end{aligned},
	\label{eq26}
\end{equation}
where ${{{\bf{B}}}(s)}$ contains only variable $s$ and {\bf{J}} can be written as
\begin{equation}
	\begin{aligned}
{\bf{J}} = (1,f,{f^2},{f^3},{f^4},{f^5})^T
	\end{aligned}.
	\label{eq27}
\end{equation}

We get four polynomial equations according to~\eqref{eq26} and {\bf{J}} is $6 \times 1$ vector of monomials in $f$. To make the number of equations equal to the number of monomials, we multiply the first equation and the second equation of~\eqref{eq23} by ${f}$. We can get six equations, which can be written as

\begin{equation}
\begin{aligned}
 \begin{array}{l}
{\bf{C}}{\bf{X}}_3 = 0,
\end{array} 
\end{aligned}
\label{eq28}
\end{equation}
where ${{\bf{C}}}$ is a $6 \times 42$ coefficient matrix.~\eqref{eq28} can be expressed as
\begin{equation}
{{\bf{B}}'}(s){\bf{J}} = 0.
 \label{eq28_1}
\end{equation}
The highest degree of ${s}$ in matrix ${\bf{{\bf{B}}'}}(s)$ is 6.~\eqref{eq28_1} can be rewritten as
\begin{equation}
	\begin{aligned}
({s^6}{{\bf{B}}_6'}\!+\!{s^5}{{\bf{B}}_5'}\!+\!{s^4}{{\bf{B}}_4'}\!+\!{s^3}{{\bf{B}}_3'}\!+\!{s^2}{{\bf{B}}_2'}\! +\!s{{\bf{B}}_1'}\!+\!{{\bf{B}}_0'}){{\bf{J}}} = 0,
	\end{aligned}
 \label{eq29}
\end{equation}
where matrix ${{\bf{B}'}_{6}}$, ${{\bf{B}'}_{5}}$,...,${{\bf{B}'}_{0}}$ are ${6 \times 6}$ matrices. we define ${\bf{D}}$, ${\bf{N}}$, and ${\bf{L}}$ as follows

\begin{equation}
	\begin{aligned}
\begin{array}{l}
{\bf{D}} = \left[ {\begin{array}{*{20}{c}}
{\bf{0}}&{\bf{I}}& \ddots  &{\bf{0}}\\
{\bf{0}}&{\bf{0}}& \ddots &{\bf{0}}\\
 \vdots & \vdots & \ddots &{\bf{I}}\\
{ - {{\bf{B}}_0'}}&{ - {{\bf{B}}_1'}}& \ddots &{ - {{\bf{B}}_5'}}
\end{array}} \right],
{\bf{N}} = \left[ {\begin{array}{*{20}{c}}
{\bf{I}}&{\bf{0}}& \ddots &{\bf{0}}\\
{\bf{0}}&{\bf{I}}& \ddots &{\bf{0}}\\
 \vdots & \vdots & \ddots &{\bf{0}}\\
{\bf{0}}&{\bf{0}}& \ddots &{{{\bf{B}}_6'}}
\end{array}} \right],
{\bf{L}} = \left[ {\begin{array}{*{20}{c}}
{\bf{J}}\\
{s{\bf{J}}}\\
\vdots \\
{{s^{5}}{\bf{J}}}
\end{array}} \right].
\end{array}
	\end{aligned}
	\label{eq31}
\end{equation}

~\eqref{eq29} can be rewritten as ${\bf{DL}} = ${s}${\bf{NL}}$. It is easy to draw a conclusion that ${s}$ is the eigenvalue of matrix ${\bf{G}}={\bf{N}}^{-1}{\bf{D}}$.

The size of matrix ${\bf{G}}$ is $36 \times 36$. The eigenvalues of matrix ${\bf{G}}$ can be estimated from Schur decomposition. The focal length ${f}$ can be obtained from the corresponding feature vector  ${\bf{L}}$. In practical applications, according to the internal properties of vector ${\bf{L}}$, some invalid solutions can be removed. The internal properties of vector ${\bf{L}}$ include (1) ${\bf{L_{(\rm2,:)}}}>0$; (2) ${{\bf{L}}_{({\rm{3}},:)}} = {\bf{L}}_{({\rm{2}},:)}^{\rm{2}}$. The translation vector ${{\bf{\tilde t}}}$ can be extracted from the null space of the matrix ${\bf{M}}$ after when ${f},{s}$ are obtained.

\section{Experiments}
In this section, We evaluate the performance of the proposed method on both synthetic and real-world data. The proposed method is named \texttt{OUR-2AC-3DOF-f}. We select three state-of-the-art methods as baselines for comparison.

$ \bullet $\texttt{Ding-4PC-3DOF-f}~\cite{ding_f}: The method uses 4 point correspondences to calculate relative pose and focal length when the direction of gravity is known.

$ \bullet $\texttt{Bara-2AC-5DOF-f}~\cite{2ac}: The method uses 2 affine correspondences to calculate relative pose and focal length.

$ \bullet $\texttt{Kuke-6PC-5DOF-f}~\cite{Polynomial}: The method uses 6 point correspondences to calculate relative pose and focal length.

Methods \texttt{Bara-2AC-5DOF-f} and \texttt{Kuke-6PC-5DOF-f} are approaches for estimating the 5DOF relative pose and focal length, and their computations are not affected by IMU data.

We use angle difference to evaluate the accuracy of the rotation matrix and translation vector. 

$ \bullet $ Rotation error ${\varepsilon _{\bf{R}}} = \arccos (\frac{{trace({{\bf{R}}_{gt}}{{\bf{R}}^{{T}}}) - 1}}{2})$,

$ \bullet $ Translation error ${\varepsilon _{\bf{t}}} = \arccos (\frac{{{\bf{t}}_{gt}{\bf{t}}^{{T}}}}{{\left\| {{\bf{t}}_{gt}} \right\| \cdot \left\| {\bf{t}} \right\|}})$,

$ \bullet $ Focal length error ${\varepsilon _f} = \frac{{\left| {{f_{gt}} - f} \right|}}{{{f_{gt}}}}$,\\
where ${{\bf{R}}_{gt}}$, ${{\bf{t}}_{gt}}$ and ${{{f}}_{gt}}$ are the ground truth rotation matrix, translation vector and focal length, respectively. ${\bf{R}}$, ${\bf{t}}$ and ${{f}}$ are the estimated value.

\subsection{Experiments on synthetic data}
To evaluate the performance of the proposed method and state-of-art method, we generate 100 3D points on a 3D cube in which the range of values for the X-axis and Y-axis are in the range of -5 to 5 meters, and the Z-axis is in the range of 5 to 20 meters. The resolution of the camera is $640 \times 480$ pixels. The principal point is (320,240) pixels. We randomly select the camera focal length to $f \in [100,1000]$. The rotation angle between two neighboring views ranges from -10° to 10°. During the experiment, image noise, roll angle noise, and pitch angle noise are added, respectively. We add noise to the affine matrix according to the method in~\cite{barath_homography}. The affine matrix can be represented by first-order approximations of the homography matrix, which can be calculated by four image points. The relation between the affine matrix and the homography matrix can be expressed as
\begin{equation}
	\begin{aligned}
          	{{a_{11}} = \frac{{{h_{11}} - {h_{31}}{u_j }}}{b}},{\qquad \quad}&{{a_{21}} = \frac{{{h_{21}} - {h_{31}}{v_j }}}{b}},\\
     	{{a_{12}}= \frac{{{h_{12}} - {h_{32}}{u_j }}}{b}},{\qquad \quad}&{{a_{22}} =    \frac{{{h_{22}} - {h_{32}}{v_j }}}{b}},
	\end{aligned}
\end{equation}
where $b = {u_i}{h_{31}} + {v_i}{h_{32}} + {h_{33}}$, and ${h_{ij}}$ represents the ${i}$-th row and the ${j}$-th column of the homography matrix ${\bf{H}}$. 
\subsubsection{Number stability}
 We define the numerical stability of the focal length error as ${\xi _{{f}}} = {{\left| {{f_{gt}} - f} \right|} \mathord{\left/ {\vphantom {{\left| {{f_{gt}} - f} \right|} {{f_{gt}}}}} \right. \kern-\nulldelimiterspace} {{f_{gt}}}}$, the rotation error as ${\xi _{\bf{R}}} = {\left\| {{{\bf{R}}_{gt}} - {\bf{R}}} \right\|_F}$, and the translation error as ${\xi _{\bf{t}}} = {\left\| {{{\bf{t}}_{gt}}/{\rm{norm}}({{\bf{t}}_{gt}}) - {\bf{t}}/{\rm{norm}}({\bf{t}})} \right\|_2}$. Each method is executed 1000 times on noise-free data.
\begin{figure}[htbp]
  \centering
     \subfloat[]{
     \begin{minipage}[t]{0.33\linewidth}
     \centering
     \includegraphics[width=0.9\linewidth]{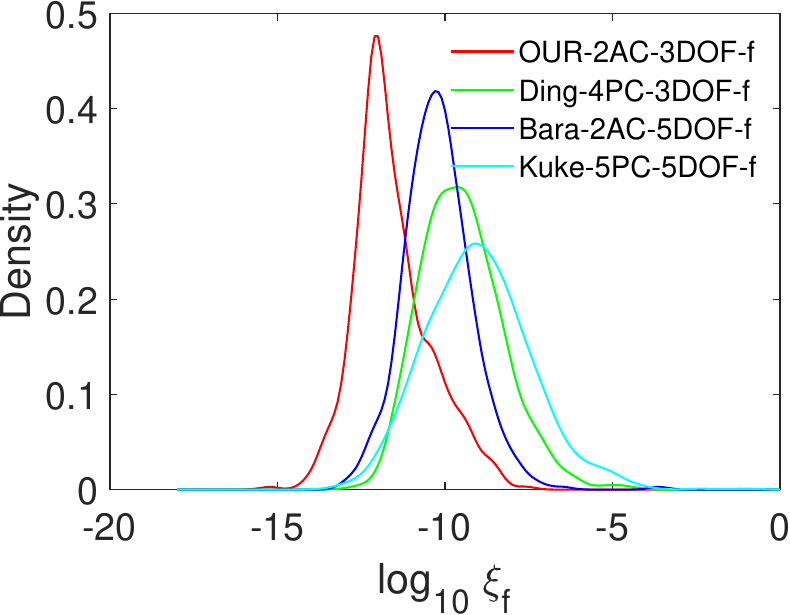}
     \end{minipage}%
      }%
      \subfloat[]{
     \begin{minipage}[t]{0.33\linewidth}
     \centering
     \includegraphics[width=0.9\linewidth]{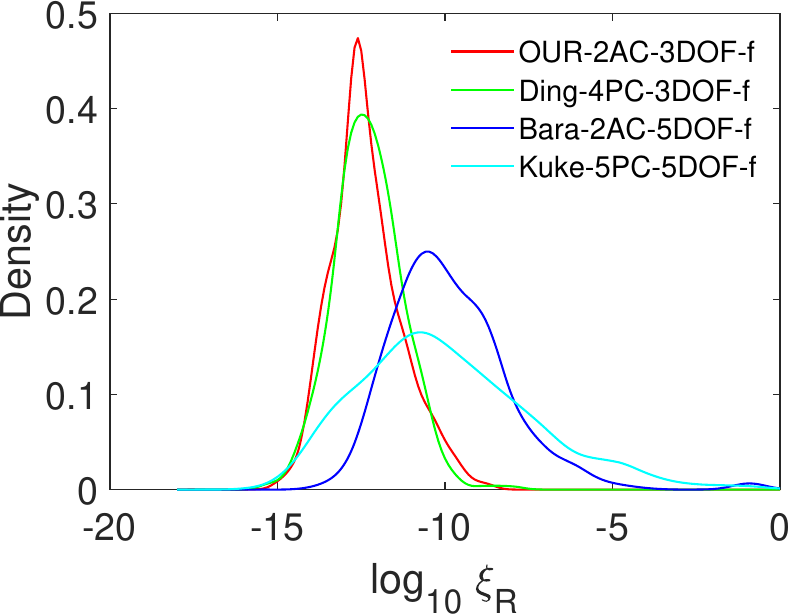}
     \end{minipage}%
      }%
      \subfloat[]{
     \begin{minipage}[t]{0.33\linewidth}
     \centering
     \includegraphics[width=0.9\linewidth]{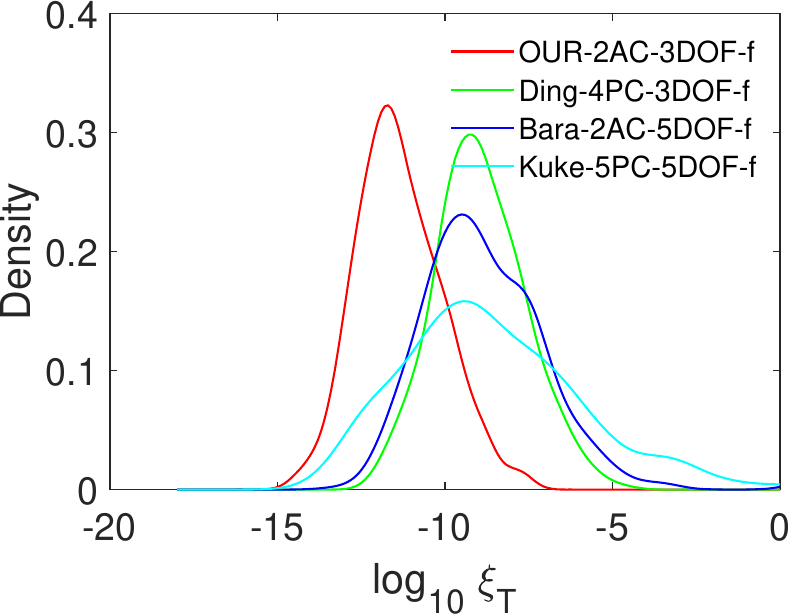}
     \end{minipage}%
      }%
    \centering
    \caption{ Probability density functions over focal length, rotation error and translation error noise free for 1000 runs under random motion. (a) Focal length error, (b) Rotation error, (b) Translation error.}
    \label{fig3}
\end{figure}
\vspace{-8pt} 

\begin{table}[http]
\centering
\caption{Numerical stability.}.
\vspace{-8pt} 
\begin{tabular}{ccccc}%
\hline  
    & ${\xi _{{f}}}$   & ${\xi _{\bf{R}}}$ & ${\xi _{\bf{t}}}$  \\ 
\hline  
 \texttt{OUR-2AC-3DOF-f} & \textbf{1.5579e-12}           & \textbf{2.8084e-13}         & \textbf{3.4358e-12}         \\
 \texttt{Ding-4PC-3DOF-f} & 2.5986e-10	& 4.9832e-13	& 7.4517e-10 \\
 \texttt{Bara-2AC-5DOF-f} & 5.4426e-11	& 7.8316e-11    & 9.1271e-10 \\
 \texttt{Kuke-6PC-5DOF-f} & 8.4889e-10	& 3.9995e-11    & 9.2171e-12\\
\hline  
\end{tabular}
\label{tab1}
\end{table}

 Fig.~\ref{fig3} shows the numerical stability of the four solvers in relative pose estimation with unknown focal length. Fig.~\ref{fig3} displays the density distribution across the logarithmic errors (base 10) of the focal length, rotation, and translation. Table~\ref{tab1} shows the median numerical stability from 1000 runs, with the minimum values denoted in black. The following conclusions can be drawn: Method \texttt{OUR-2AC-3DOF-f} significantly outperforms the three baseline methods, achieving lower errors in focal length, rotation, and translation estimation. Method \texttt{OUR-2AC-3DOF-f} achieves focal length errors 2 orders of magnitude lower than \texttt{Ding-4PC-3DOF-f} and \texttt{Kuke-6PC-5DOF-f}, and 1 order of magnitude lower than \texttt{Bara-2AC-5DOF-f}. Method \texttt{OUR-2AC-3DOF-f} achieves rotation matrix errors 2 orders of magnitude lower than both \texttt{Bara-2AC-5DOF-f} and \texttt{Kuke-6PC-5DOF-f}. Method \texttt{OUR-2AC-3DOF-f} achieves translation vector errors 2 orders of magnitude smaller than those of \texttt{Ding-4PC-3DOF-f} and \texttt{Bara-2AC-5DOF-f}. In summary, the proposed method demonstrates superior numerical stability for relative pose estimation compared to the baseline methods.

\begin{figure}[tbp]
  \centering
     \subfloat[]{
     \begin{minipage}[t]{0.33\linewidth}
     \centering
     \includegraphics[width=0.99\linewidth]{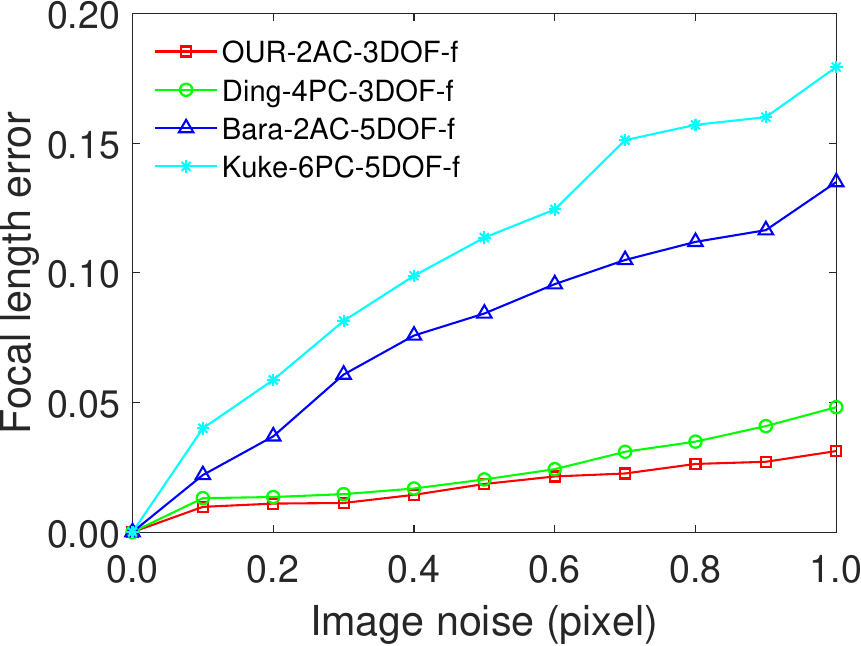}
     \end{minipage}%
      }%
      \subfloat[]{
     \begin{minipage}[t]{0.33\linewidth}
     \centering
     \includegraphics[width=0.99\linewidth]{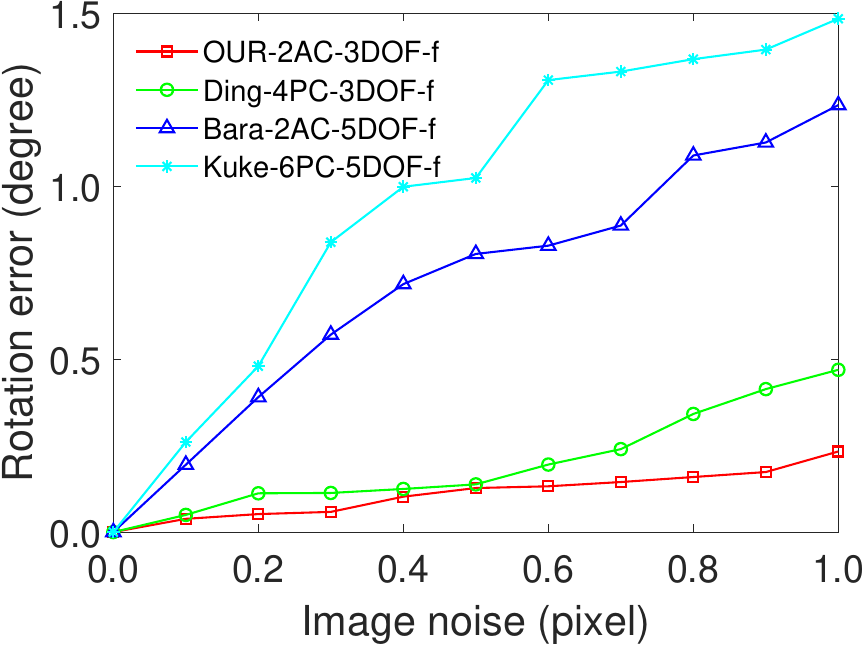}
     \end{minipage}%
      }%
      \subfloat[]{
     \begin{minipage}[t]{0.33\linewidth}
     \centering
     \includegraphics[width=0.99\linewidth]{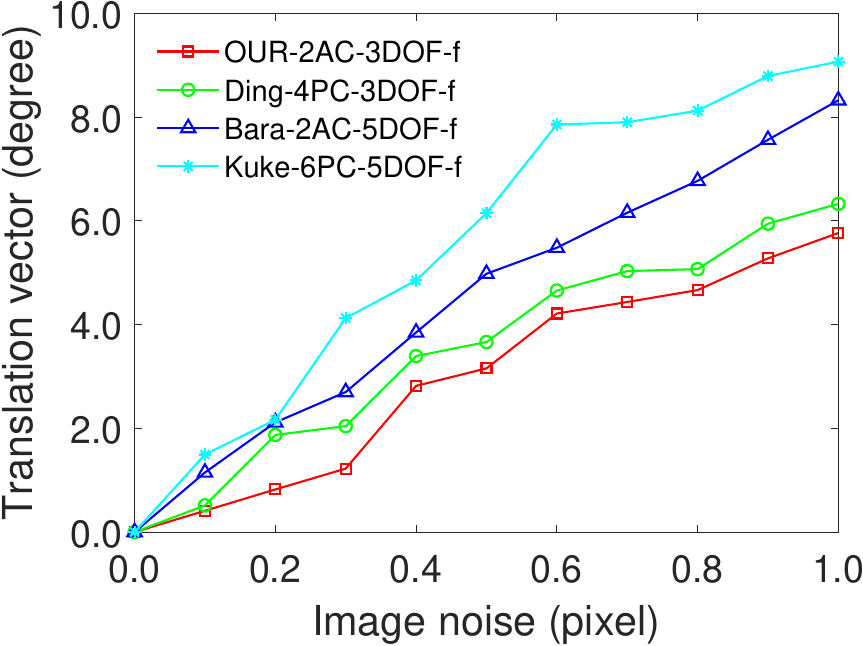}
     \end{minipage}%
      }%
    \centering
\vspace{-8pt}    
     \subfloat[]{
     \begin{minipage}[t]{0.33\linewidth}
     \centering
     \includegraphics[width=0.99\linewidth]{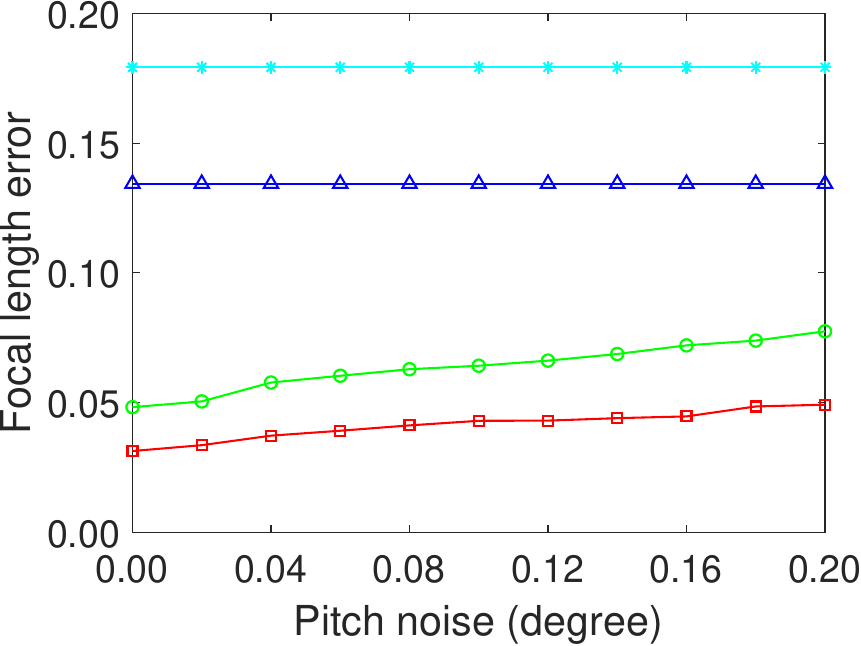}
     \end{minipage}%
      }%
      \subfloat[]{
     \begin{minipage}[t]{0.33\linewidth}
     \centering
     \includegraphics[width=0.99\linewidth]{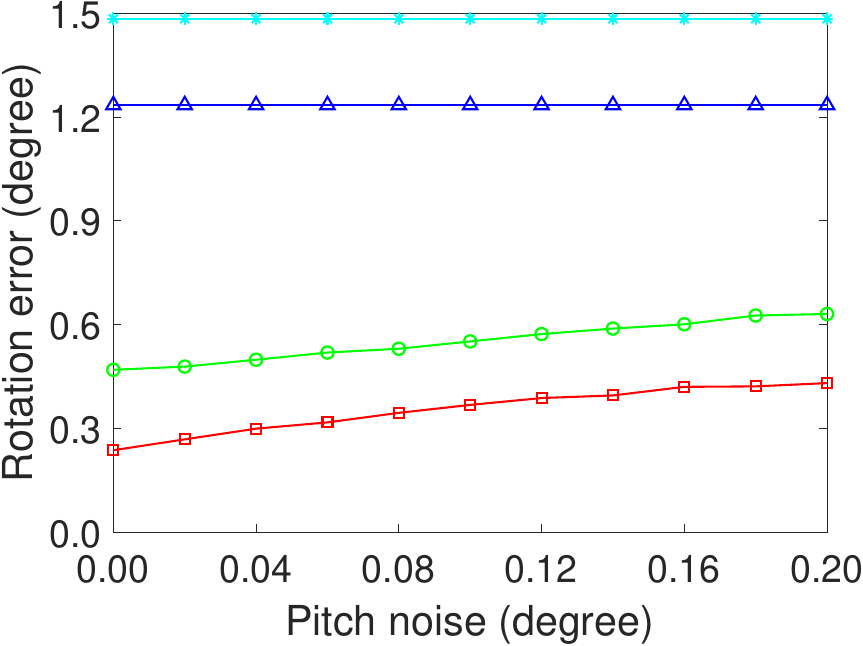}
     \end{minipage}%
      }%
      \subfloat[]{
     \begin{minipage}[t]{0.33\linewidth}
     \centering
     \includegraphics[width=0.99\linewidth]{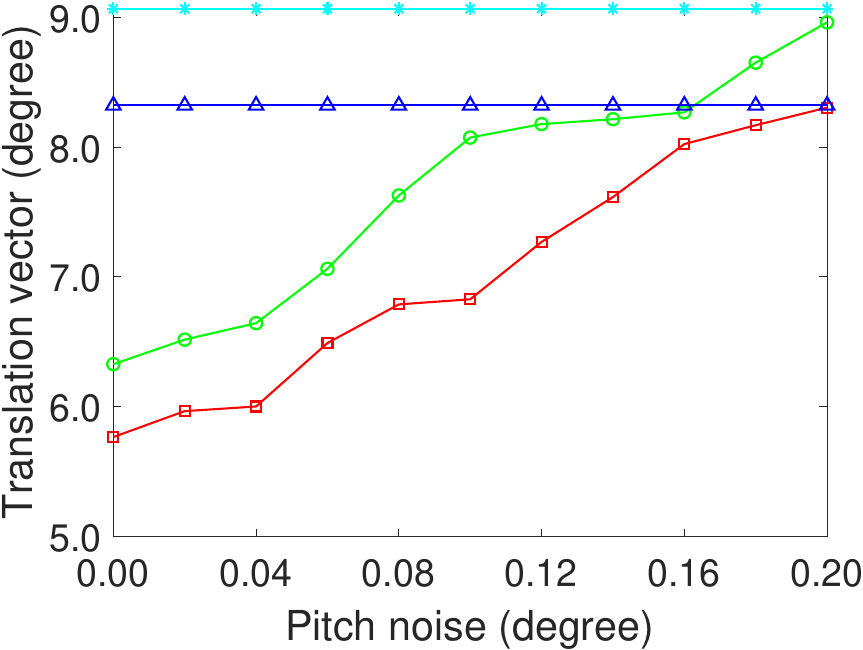}
     \end{minipage}%
      }%
    \centering
\vspace{-8pt}    
     \subfloat[]{
     \begin{minipage}[t]{0.33\linewidth}
     \centering
     \includegraphics[width=0.99\linewidth]{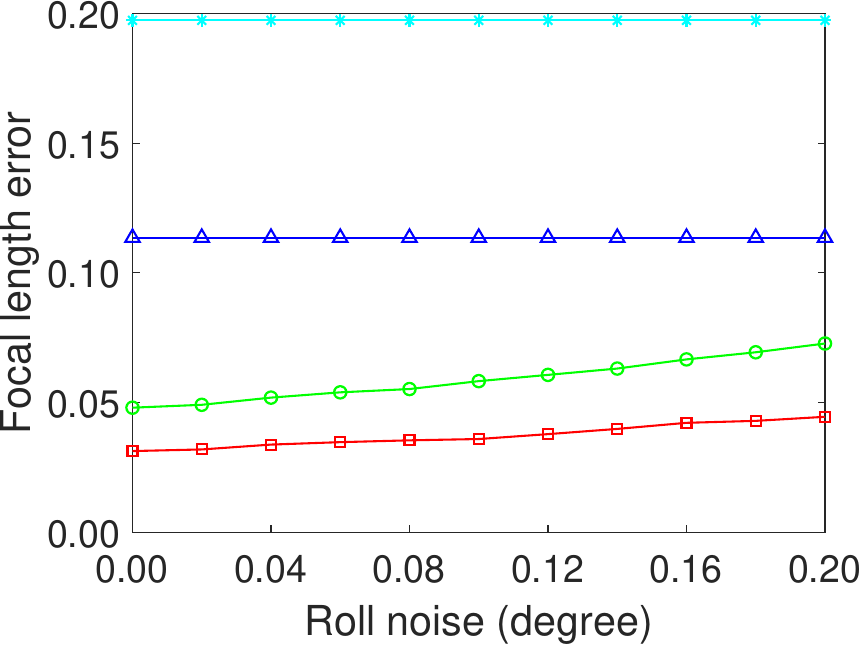}
     \end{minipage}%
      }%
      \subfloat[]{
     \begin{minipage}[t]{0.33\linewidth}
     \centering
     \includegraphics[width=0.99\linewidth]{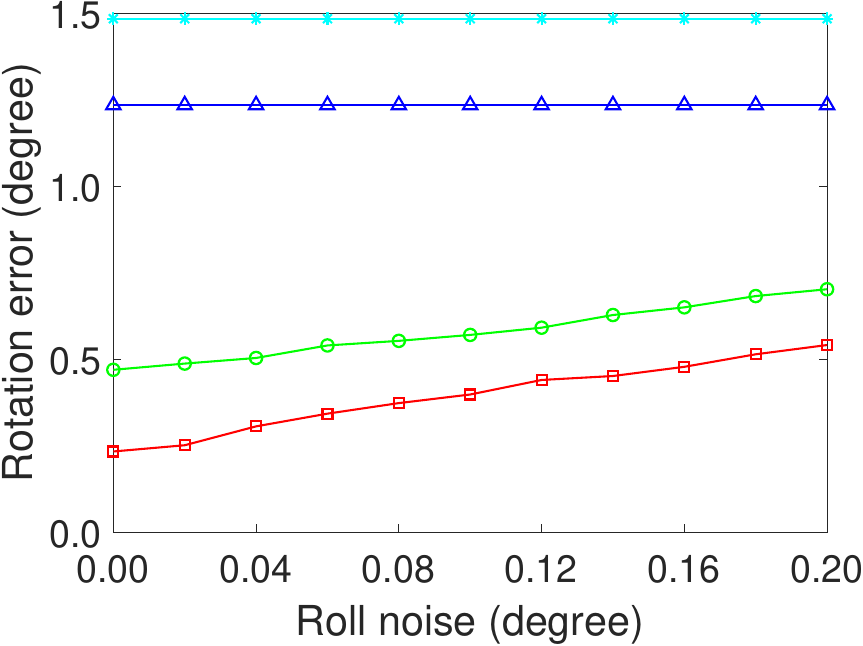}
     \end{minipage}%
      }%
      \subfloat[]{
     \begin{minipage}[t]{0.33\linewidth}
     \centering
     \includegraphics[width=0.99\linewidth]{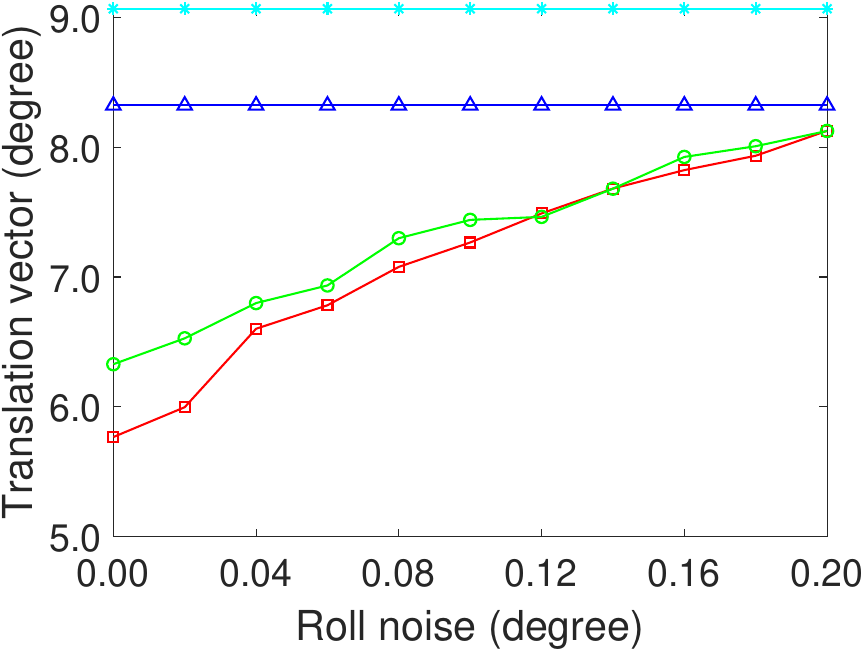}
     \end{minipage}%
      }%
    \centering 
    \caption{Focal length, Rotation and translation error for four methods under random motion. First column: Focal length error (\%); Second column: Rotation error (degree); Third column: translation error (degree): (a)(b)(c) adding image noise with perfect IMU data. (d)(e)(f) adding pitch angle noise and fix the image noise as 1.0 pixel. (g)(h)(i) adding roll angle noise and fix the image noise as 1.0 pixel.}
    \label{fig4}
\end{figure}

\subsubsection{Noise Resilience}
This section examines the performance of all four methods under image noise  and IMU noise. The evaluation covers four distinct motion patterns: random motion (${\bf{t}} =[{\begin{array}{*{20}{c}}{{t_x}}&{{t_y}}&{{t_z}}\end{array}}]^{{T}}$), planar motion (${\bf{t}} =[{\begin{array}{*{20}{c}}{{t_x}}&{{0}}&{{t_z}}\end{array}}]^{{T}}$), sideways motion (${\bf{t}} =[{\begin{array}{*{20}{c}}{{t_x}}&{{0}}&{{0}}\end{array}}]^{{T}}$), forward motion (${\bf{t}} =[{\begin{array}{*{20}{c}}{{0}}&{{0}}&{{t_z}}\end{array}}]^{{T}}$). Image noise is one pixel in four motion modes when noise is added to the pitch angle and roll angle.

\begin{figure}[tbp]
    \centering
     \subfloat[]{
     \begin{minipage}[t]{0.33\linewidth}
     \centering
     \includegraphics[width=0.99\linewidth]{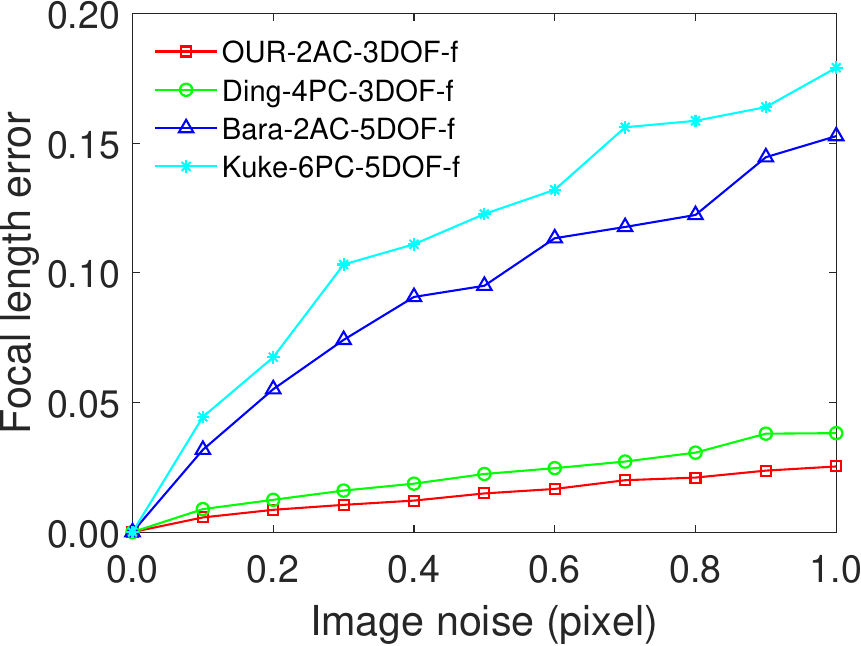}
     \end{minipage}%
      }%
      \subfloat[]{
     \begin{minipage}[t]{0.33\linewidth}
     \centering
     \includegraphics[width=0.99\linewidth]{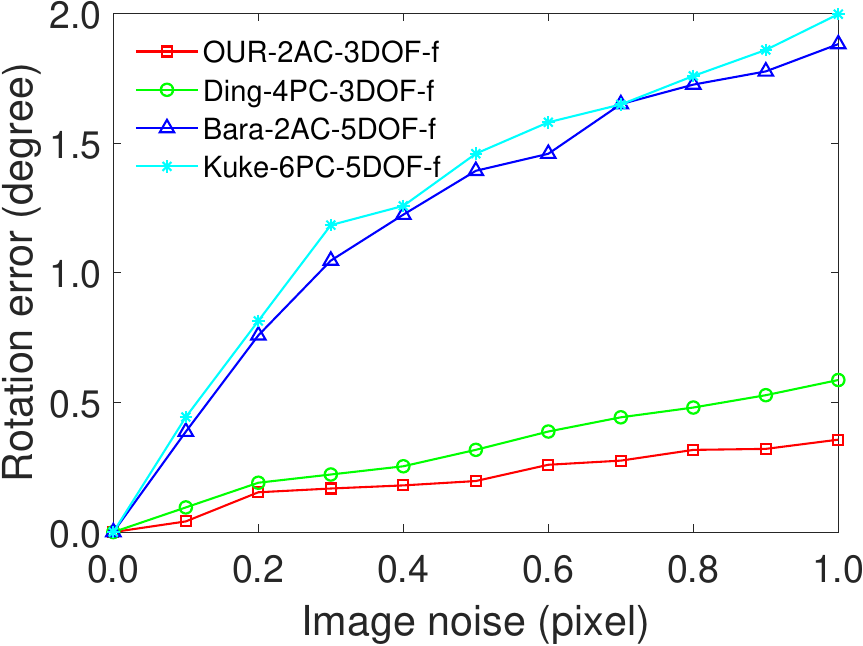}
     \end{minipage}%
      }%
      \subfloat[]{
     \begin{minipage}[t]{0.33\linewidth}
     \centering
     \includegraphics[width=0.99\linewidth]{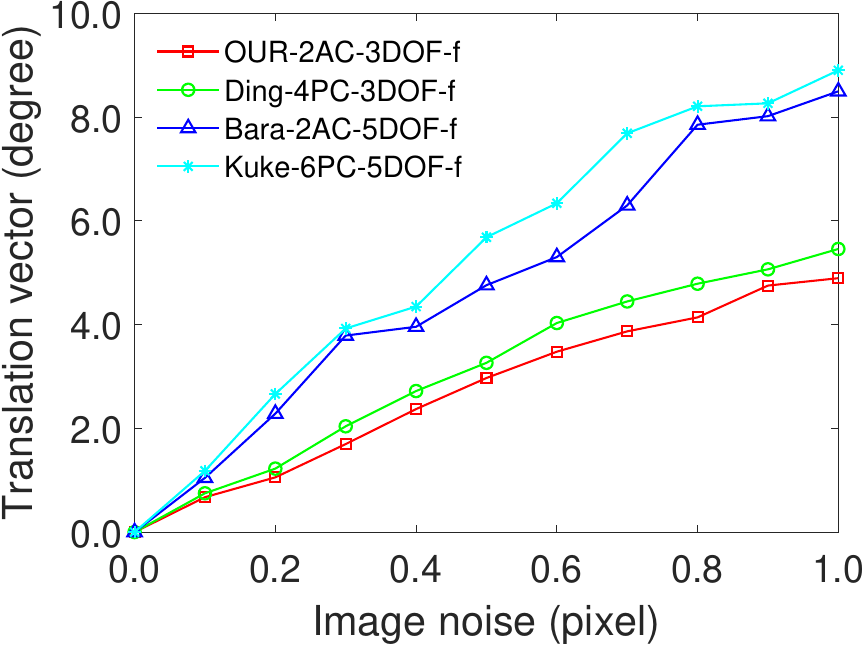}
     \end{minipage}%
      }%
    \centering
\vspace{-8pt}    
     \subfloat[]{
     \begin{minipage}[t]{0.33\linewidth}
     \centering
     \includegraphics[width=0.99\linewidth]{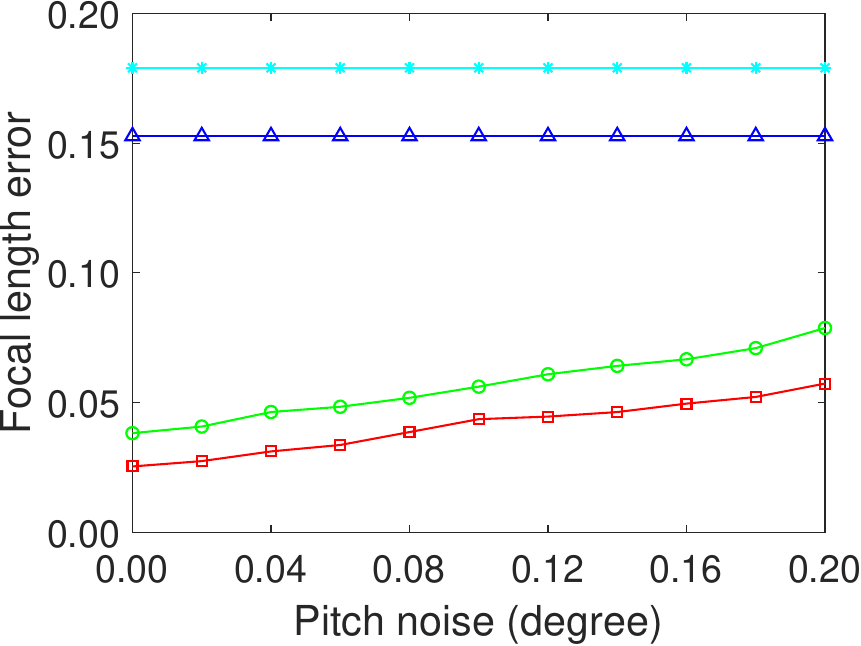}
     \end{minipage}%
      }%
      \subfloat[]{
     \begin{minipage}[t]{0.33\linewidth}
     \centering
     \includegraphics[width=0.99\linewidth]{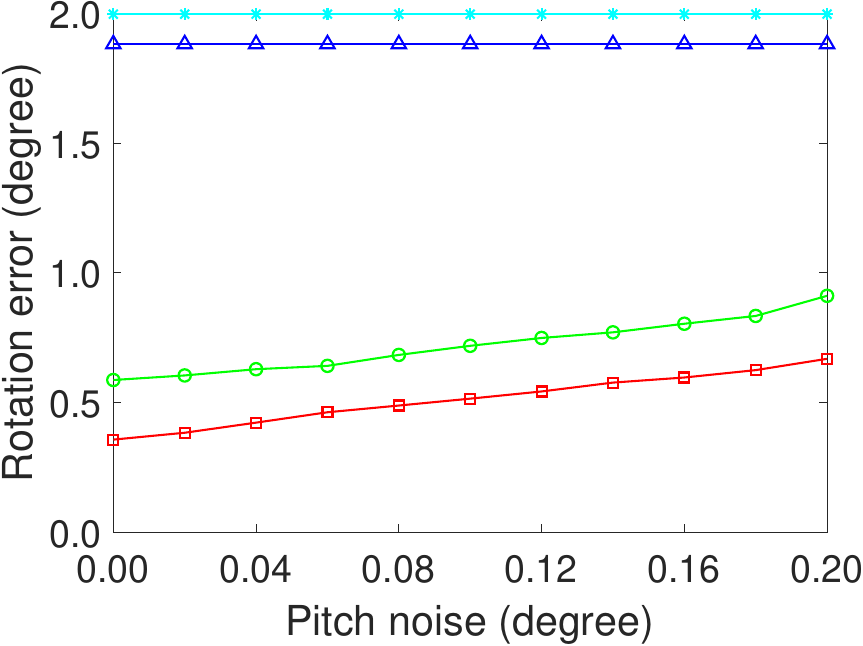}
     \end{minipage}%
      }%
      \subfloat[]{
     \begin{minipage}[t]{0.33\linewidth}
     \centering
     \includegraphics[width=0.99\linewidth]{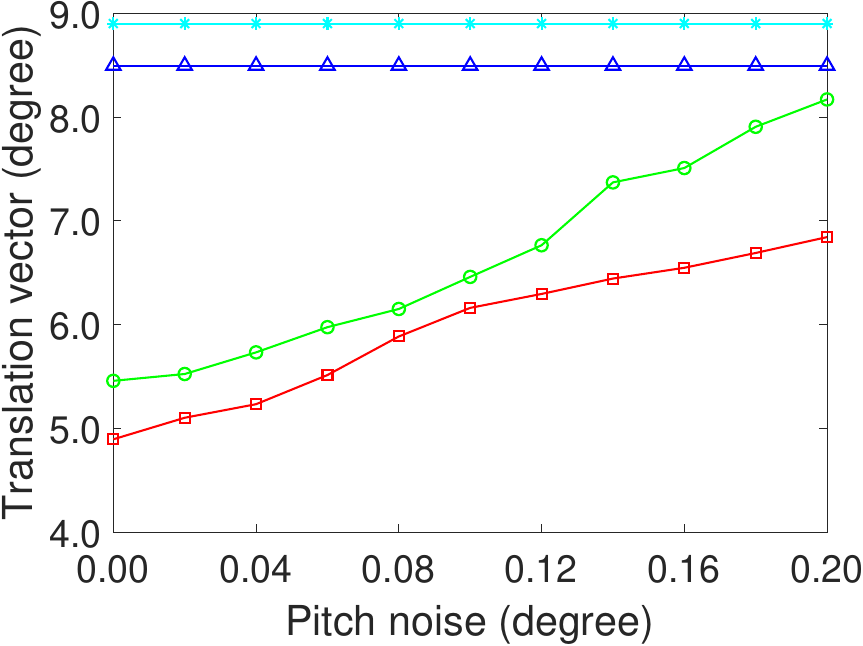}
     \end{minipage}%
      }%
    \centering
\vspace{-8pt}
     \subfloat[]{
     \begin{minipage}[t]{0.33\linewidth}
     \centering
     \includegraphics[width=0.99\linewidth]{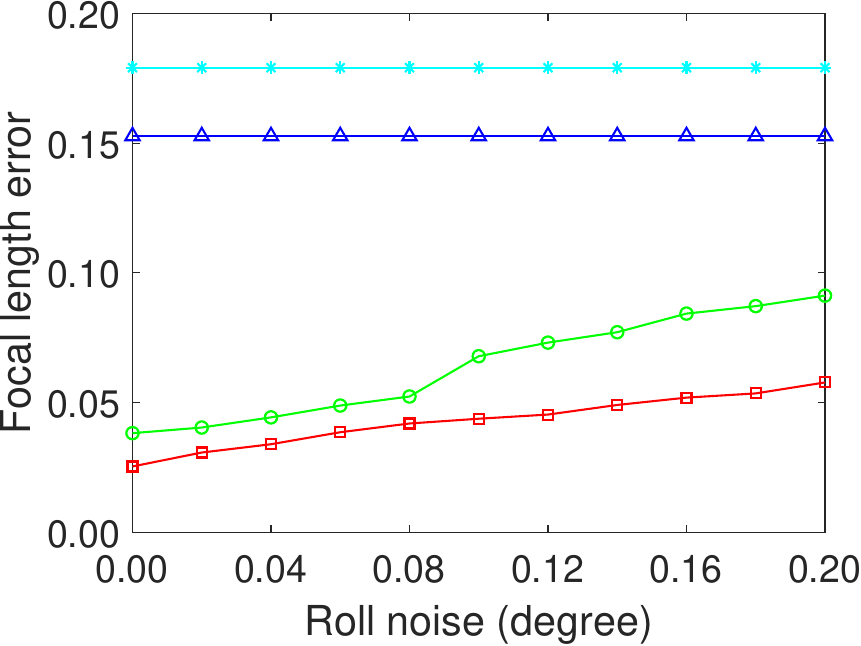}
     \end{minipage}%
      }%
      \subfloat[]{
     \begin{minipage}[t]{0.33\linewidth}
     \centering
     \includegraphics[width=0.99\linewidth]{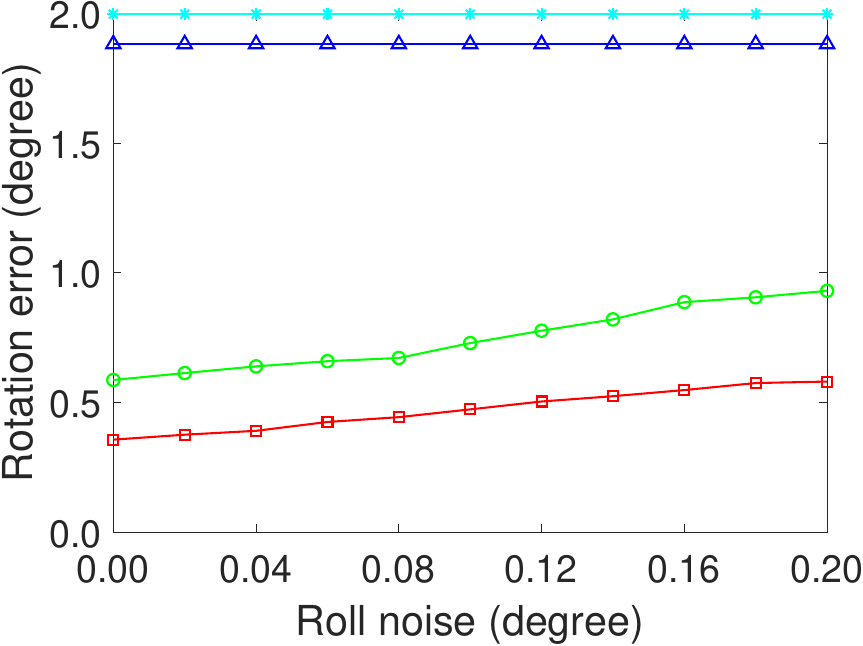}
     \end{minipage}%
      }%
      \subfloat[]{
     \begin{minipage}[t]{0.33\linewidth}
     \centering
     \includegraphics[width=0.99\linewidth]{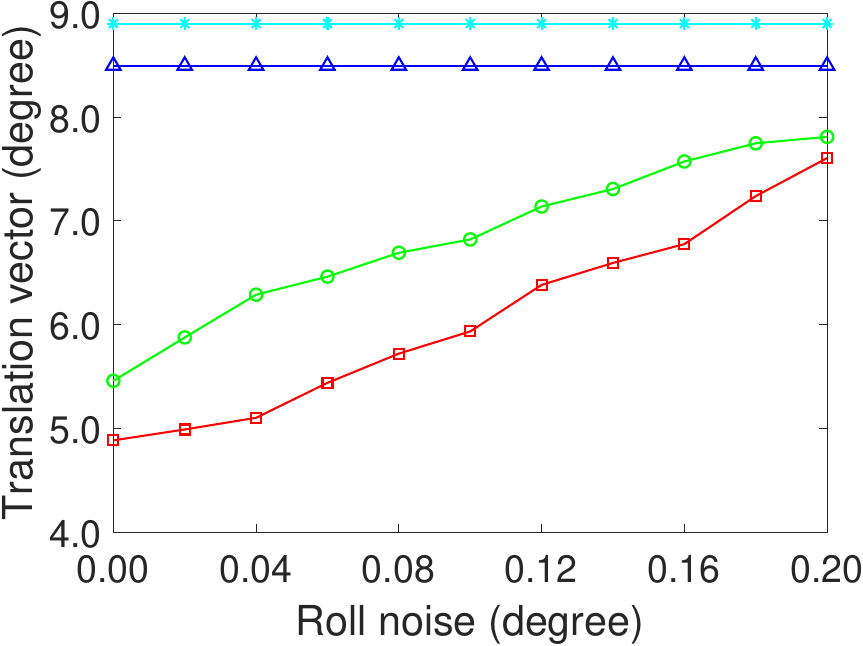}
     \end{minipage}%
      }%
    \centering 
    \caption{Focal length, Rotation and translation error for four methods under planar motion. First column: Focal length error (\%); Second column: Rotation error (degree); Third column: translation error (degree): (a)(b)(c) adding image noise with perfect IMU data. (d)(e)(f) adding pitch angle noise and fix the image noise as 1.0 pixel. (g)(h)(i) adding roll angle noise and fix the image noise as 1.0 pixel }
    \label{fig5}
\end{figure}

\begin{figure}[tbp]
    \centering
     \subfloat[]{
     \begin{minipage}[t]{0.33\linewidth}
     \centering
     \includegraphics[width=0.99\linewidth]{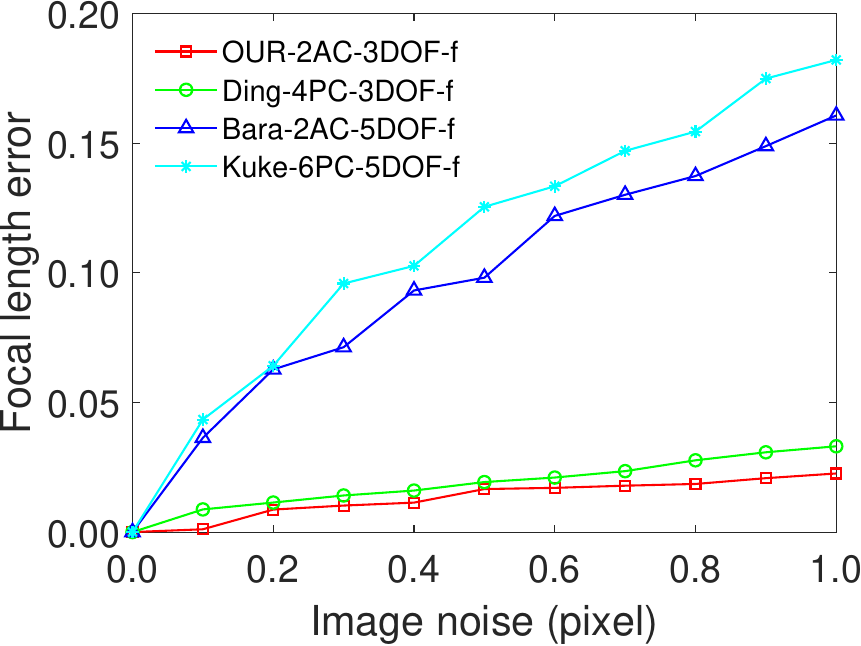}
     \end{minipage}%
      }%
      \subfloat[]{
     \begin{minipage}[t]{0.33\linewidth}
     \centering
     \includegraphics[width=0.99\linewidth]{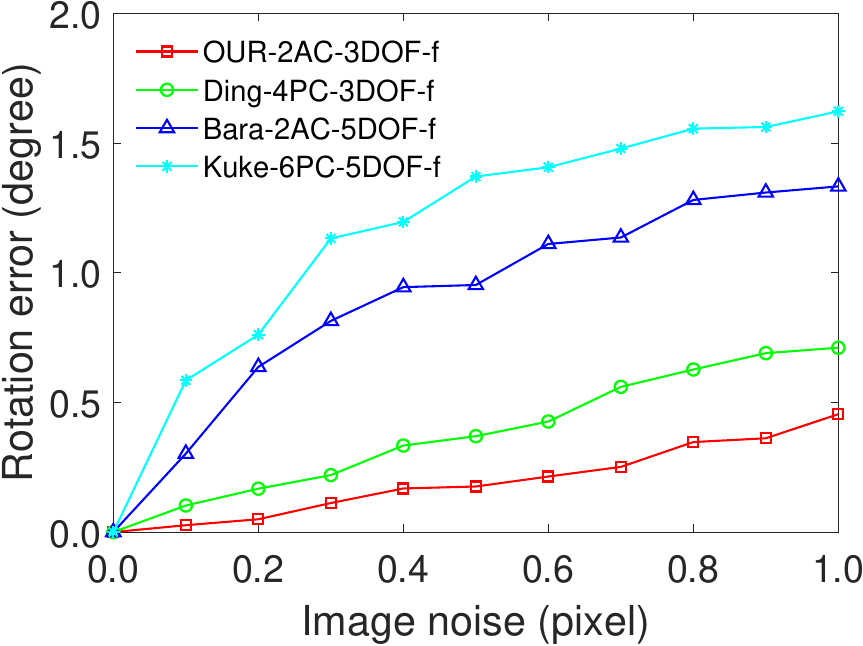}
     \end{minipage}%
      }%
      \subfloat[]{
     \begin{minipage}[t]{0.33\linewidth}
     \centering
     \includegraphics[width=0.99\linewidth]{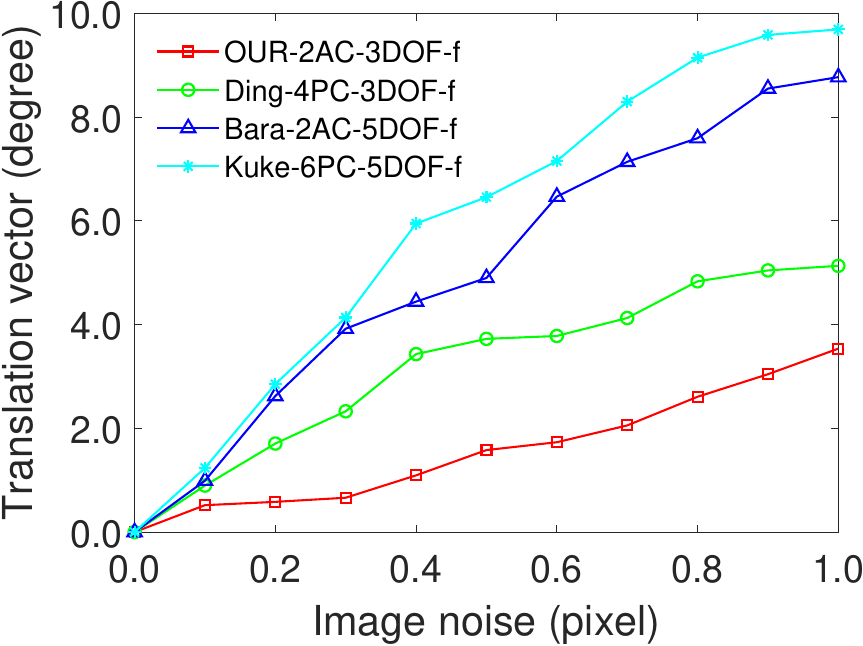}
     \end{minipage}%
      }%
    \centering
\vspace{-8pt}    
     \subfloat[]{
     \begin{minipage}[t]{0.33\linewidth}
     \centering
     \includegraphics[width=0.99\linewidth]{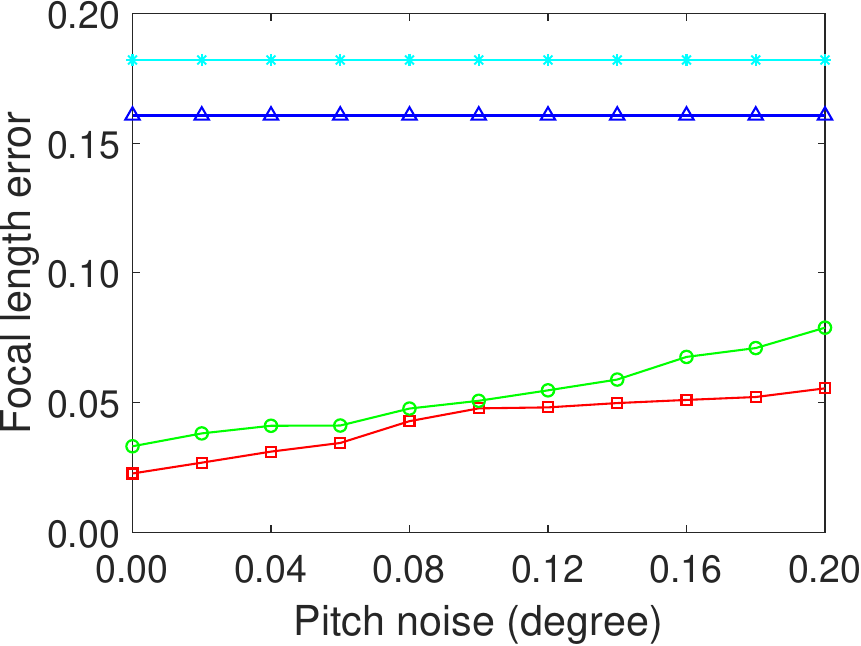}
     \end{minipage}%
      }%
      \subfloat[]{
     \begin{minipage}[t]{0.33\linewidth}
     \centering
     \includegraphics[width=0.99\linewidth]{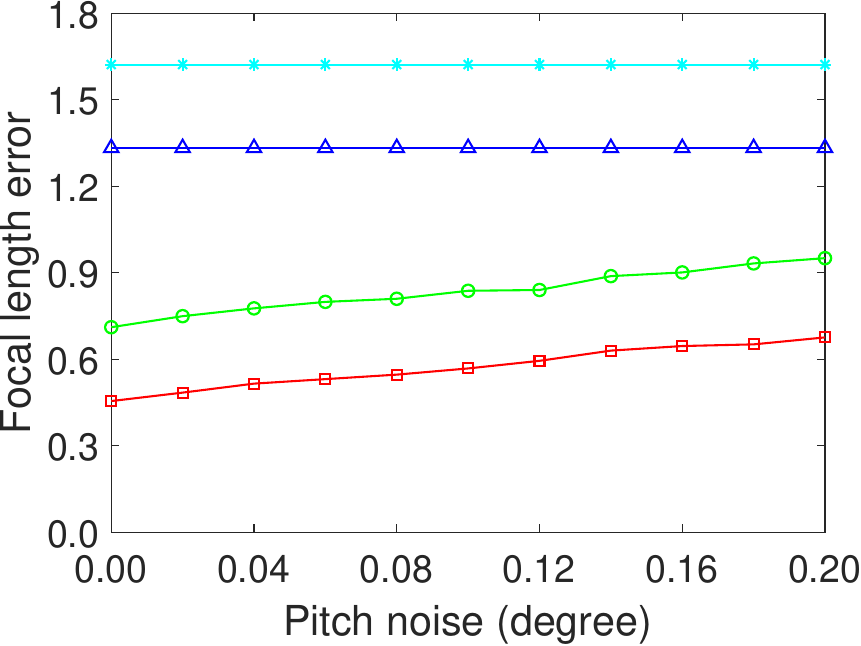}
     \end{minipage}%
      }%
      \subfloat[]{
     \begin{minipage}[t]{0.33\linewidth}
     \centering
     \includegraphics[width=0.99\linewidth]{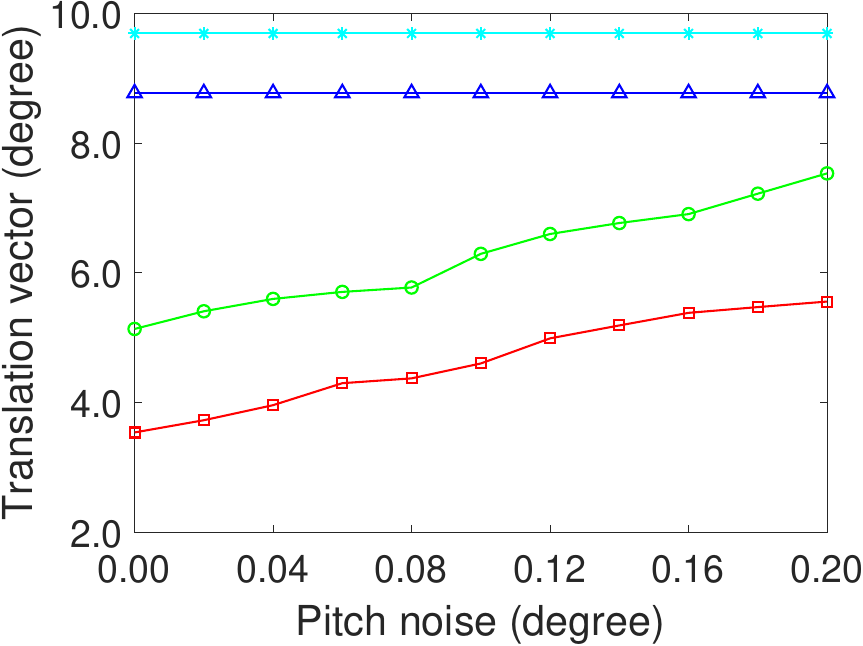}
     \end{minipage}%
      }%
    \centering
\vspace{-8pt}
     \subfloat[]{
     \begin{minipage}[t]{0.33\linewidth}
     \centering
     \includegraphics[width=0.99\linewidth]{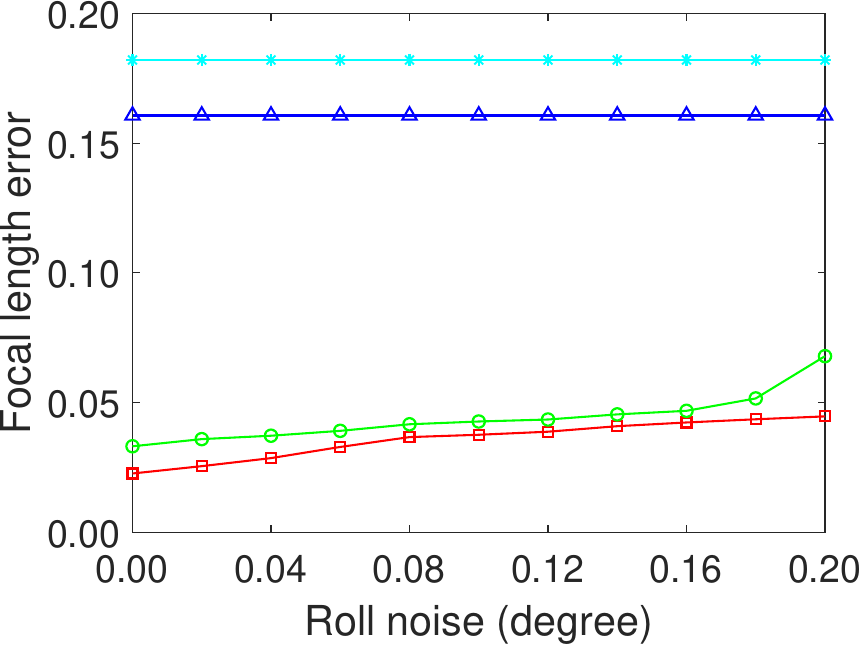}
     \end{minipage}%
      }%
      \subfloat[]{
     \begin{minipage}[t]{0.33\linewidth}
     \centering
     \includegraphics[width=0.99\linewidth]{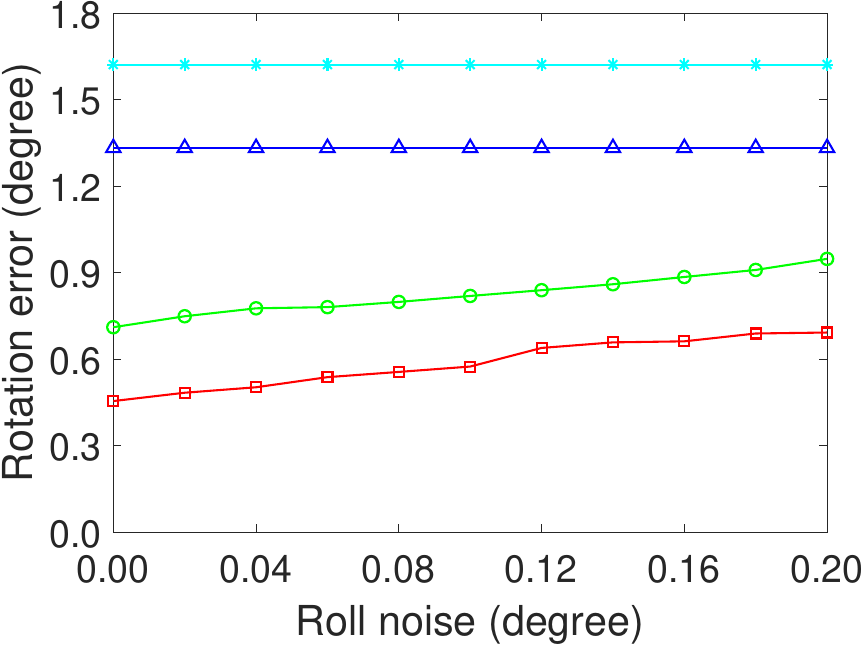}
     \end{minipage}%
      }%
      \subfloat[]{
     \begin{minipage}[t]{0.33\linewidth}
     \centering
     \includegraphics[width=0.99\linewidth]{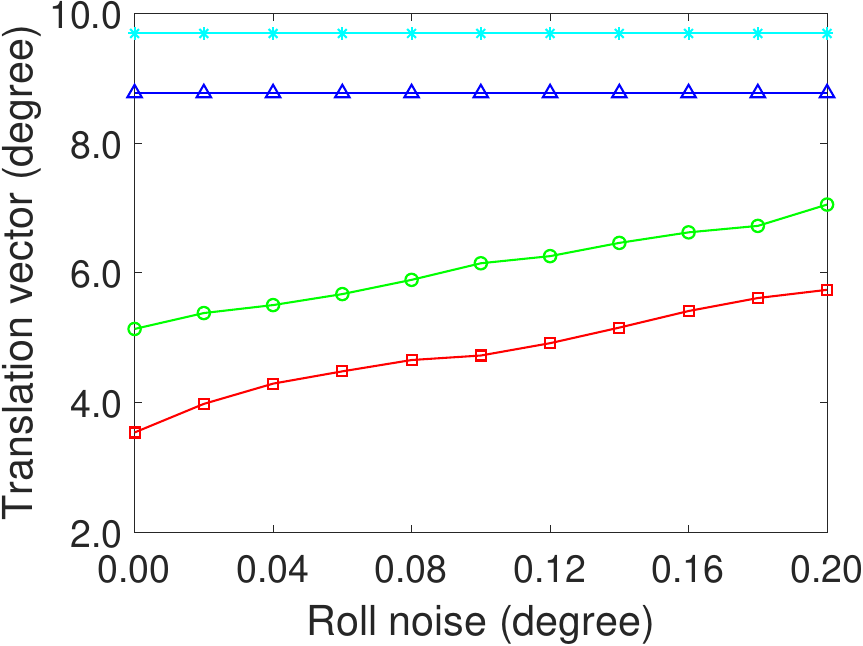}
     \end{minipage}%
      }%
    \centering 
    \caption{Focal length, Rotation and translation error for four methods under sideways motion. First column: Focal length error (\%); Second column: Rotation error (degree); Third column: translation error (degree): (a)(b)(c) adding image noise with perfect IMU data. (d)(e)(f) adding pitch angle noise and fix the image noise as 1.0 pixel. (g)(h)(i) adding roll angle noise and fix the image noise as 1.0 pixel }
    \label{fig6}
\end{figure}

\begin{figure}[ht!]
\vspace*{-0.4cm}
      \centering
     \subfloat[]{
     \begin{minipage}[t]{0.33\linewidth}
     \centering
     \includegraphics[width=0.99\linewidth]{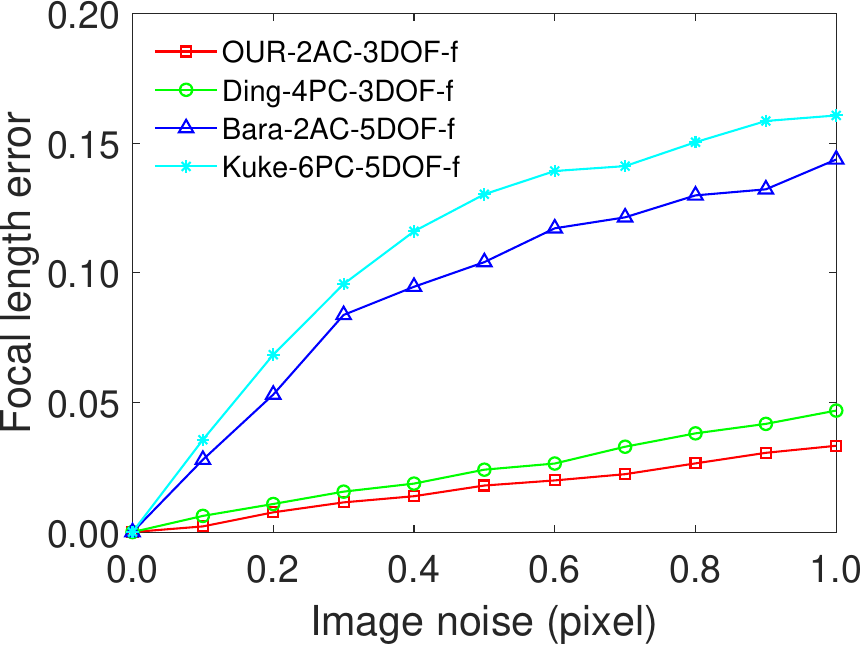}
     \end{minipage}%
      }%
      \subfloat[]{
     \begin{minipage}[t]{0.33\linewidth}
     \centering
     \includegraphics[width=0.99\linewidth]{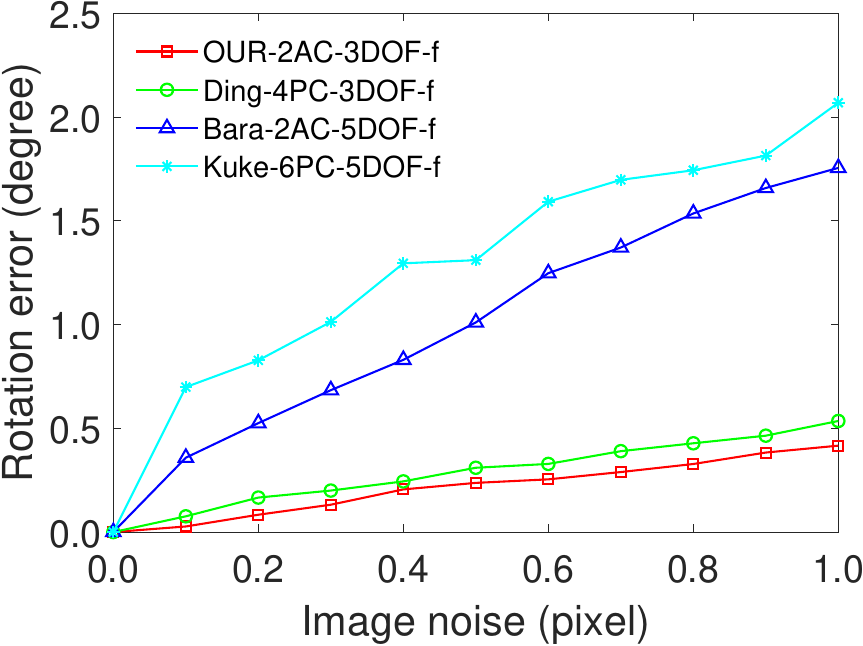}
     \end{minipage}%
      }%
      \subfloat[]{
     \begin{minipage}[t]{0.33\linewidth}
     \centering
     \includegraphics[width=0.99\linewidth]{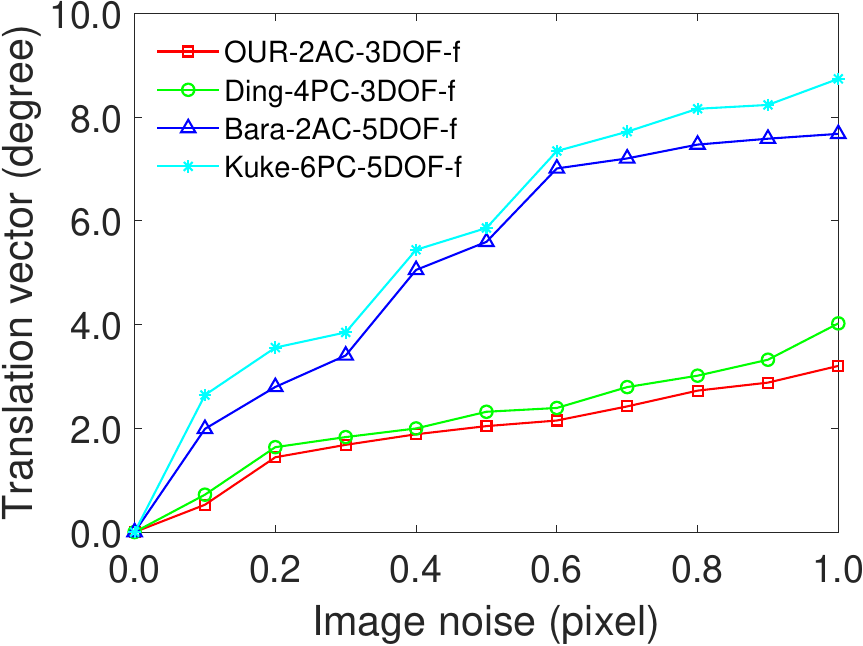}
     \end{minipage}%
      }%
    \centering
\vspace{-8pt}
    
     \subfloat[]{
     \begin{minipage}[t]{0.33\linewidth}
     \centering
     \includegraphics[width=0.99\linewidth]{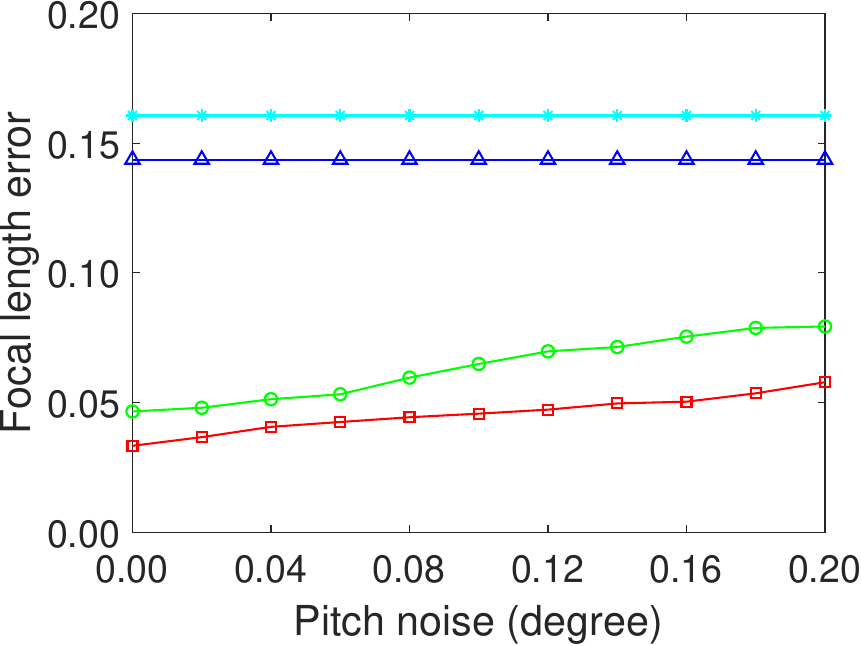}
     \end{minipage}%
      }%
      \subfloat[]{
     \begin{minipage}[t]{0.33\linewidth}
     \centering
     \includegraphics[width=0.99\linewidth]{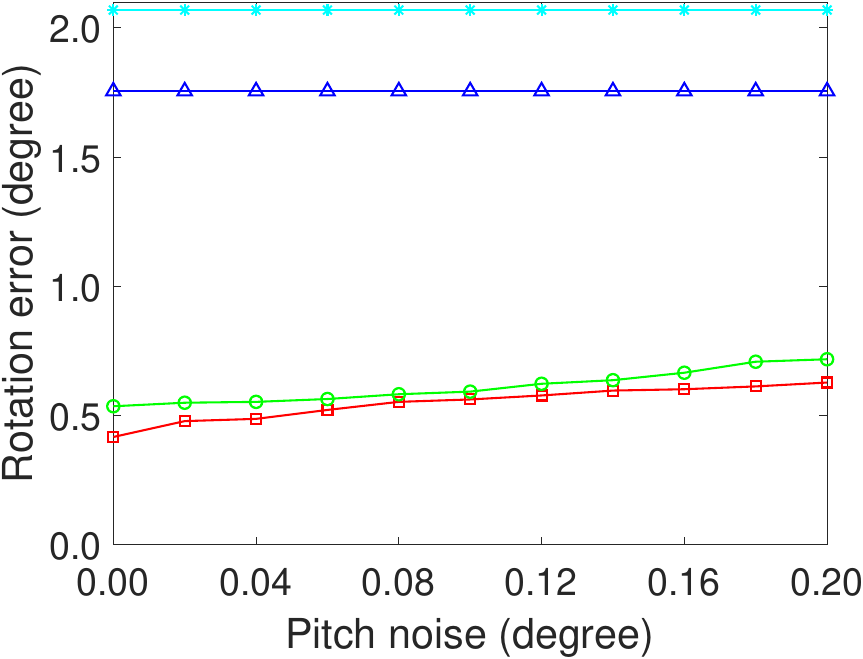}
     \end{minipage}%
      }%
      \subfloat[]{
     \begin{minipage}[t]{0.33\linewidth}
     \centering
     \includegraphics[width=0.99\linewidth]{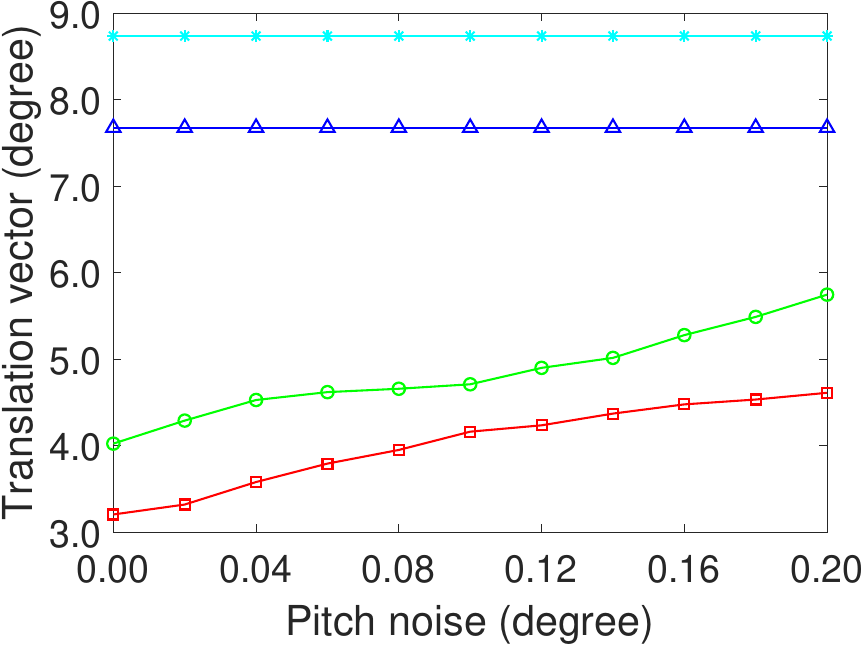}
     \end{minipage}%
      }%
    \centering
\vspace{-8pt}
    
     \subfloat[]{
     \begin{minipage}[t]{0.33\linewidth}
     \centering
     \includegraphics[width=0.99\linewidth]{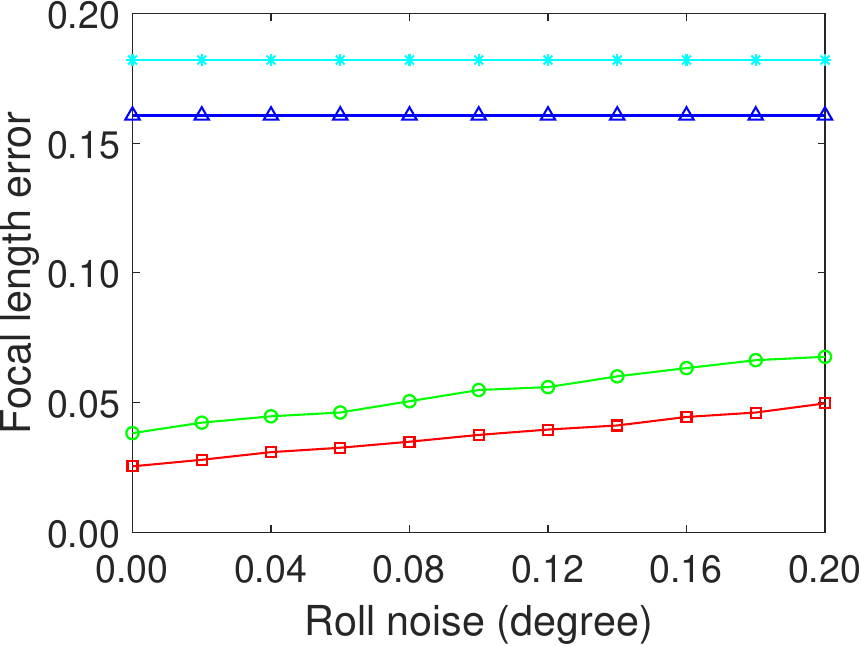}
     \end{minipage}%
      }%
      \subfloat[]{
     \begin{minipage}[t]{0.33\linewidth}
     \centering
     \includegraphics[width=0.99\linewidth]{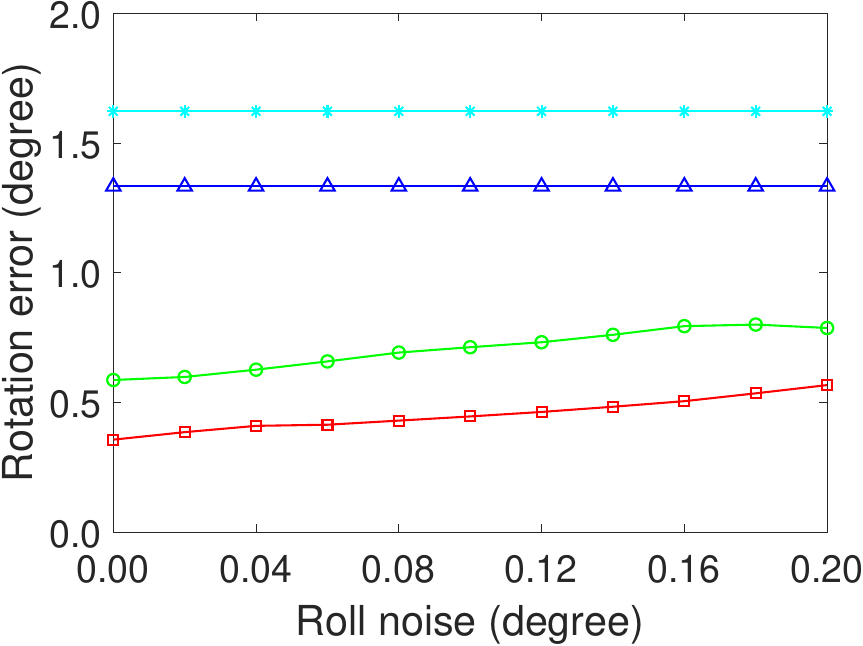}
     \end{minipage}%
      }%
      \subfloat[]{
     \begin{minipage}[t]{0.33\linewidth}
     \centering
     \includegraphics[width=0.99\linewidth]{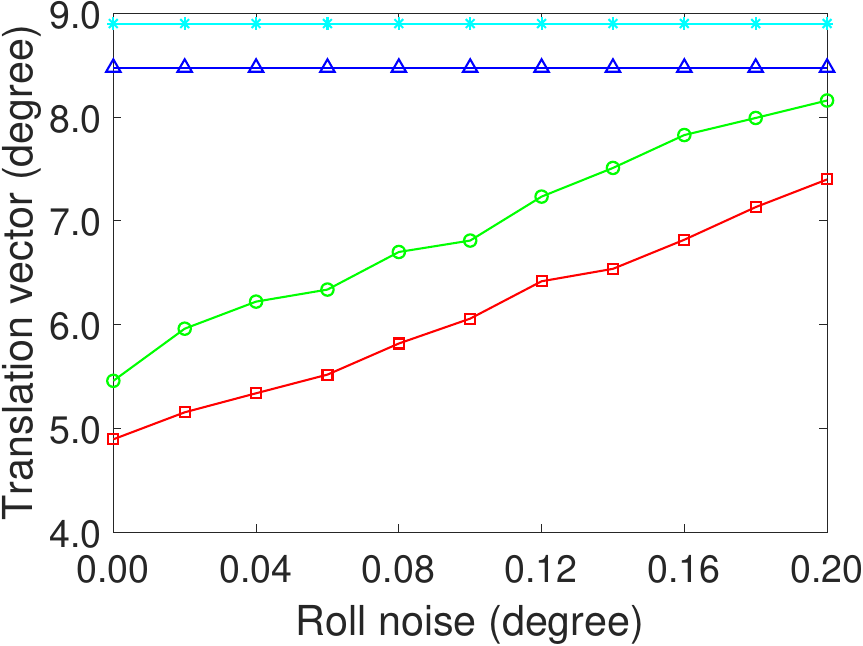}
     \end{minipage}%
      }%
    \centering
    
 \caption{Focal length, rotation and translation error for four methods under forward motion. First column: Focal length error (\%); Second column: Rotation error (degree); Third column: translation error (degree): (a)(b)(c) adding image noise with perfect IMU data. (d)(e)(f) adding pitch angle noise and fix the image noise as 1.0 pixel. (g)(h)(i) adding roll angle noise and fix the image noise as 1.0 pixel}
    \label{fig7}
\end{figure}

Figs.\ref{fig4}-\ref{fig7} present the estimation errors under four motion patterns. The three rows respectively show the impact of image noise, pitch angle noise, and roll angle noise on the estimation accuracy of focal length, rotation, and translation across all four methods. The 5-DOF methods \texttt{Bara-2AC-5DOF-f} and \texttt{Kuke-6PC-5DOF-f} are not affected by the angle noise values from the inertial measurement unit. Therefore, these two methods maintain constant estimation accuracy under increasing IMU angular noise, as evidenced in the second and third rows.

Under random motion with image noise, the proposed \texttt{OUR-2AC-3DOF-f} yields the lowest errors in focal length, rotation, and translation. It is followed by \texttt{Ding-4PC-3DOF-f}, which shows smaller errors than both \texttt{Bara-2AC-5DOF-f} and \texttt{Kuke-6PC-5DOF-f}. Between the latter two, \texttt{Bara-2AC-5DOF-f} maintains consistently lower errors than \texttt{Kuke-6PC-5DOF-f} across all metrics. When adding pitch and roll angles noise, the focal length error and rotation matrix error computed by the proposed \texttt{OUR-2AC-3DOF-f} method are both smaller than comparison methods. The proposed approach maintains superior translation accuracy in most imu noise.

Under planar motion with image noise below 0.2 pixels, the focal length and translation errors of \texttt{Bara-2AC-5DOF-f} and \texttt{Kuke-6PC-5DOF-f} remain comparable. The proposed \texttt{OUR-2AC-3DOF-f} achieves only marginally lower focal length error than \texttt{Ding-4PC-3DOF-f}, while both methods consistently outperform the other two approaches across all three error metrics. When adding pitch angle and roll angle noise, the proposed \texttt{OUR-2AC-3DOF-f} achieves lower estimation errors than \texttt{Ding-4PC-3DOF-f} in all three metrics. Furthermore, both methods maintain a consistent accuracy advantage over \texttt{Bara-2AC-5DOF-f} and \texttt{Kuke-6PC-5DOF-f} in focal length, rotation, and translation estimation.

Under sideways motion with image noise, the proposed \texttt{OUR-2AC-3DOF-f} method demonstrates superior performance across all evaluation metrics, achieving lower focal length, rotation, and translation errors than \texttt{Ding-4PC-3DOF-f}. Both methods maintain a consistent advantage over \texttt{Bara-2AC-5DOF-f} and \texttt{Kuke-6PC-5DOF-f}. Notably, under pitch and roll angle noise conditions, \texttt{OUR-2AC-3DOF-f} achieves the smallest errors among all compared methods.

In the forward motion mode, whether adding image noise or pitch angle and roll angle noise, the focal length error, rotation matrix error, and translation vector error computed by the \texttt{OUR-2AC-3DOF-f} method are all smaller than those obtained from the \texttt{Ding-4PC-3DOF-f}, \texttt{Bara-2AC-5DOF-f}, and \texttt{Kuke-6PC-5DOF-f} methods. Additionally, the errors calculated by the \texttt{Ding-4PC-3DOF-f} method in terms of focal length, rotation matrix, and translation vector are all smaller than those from the \texttt{Bara-2AC-5DOF-f} and \texttt{Kuke-6PC-5DOF-f} methods.

\subsubsection{principal point noise}
\begin{figure}[htbp]
  \centering
     \subfloat[]{
     \begin{minipage}[t]{0.33\linewidth}
     \centering
     \includegraphics[width=0.9\linewidth]{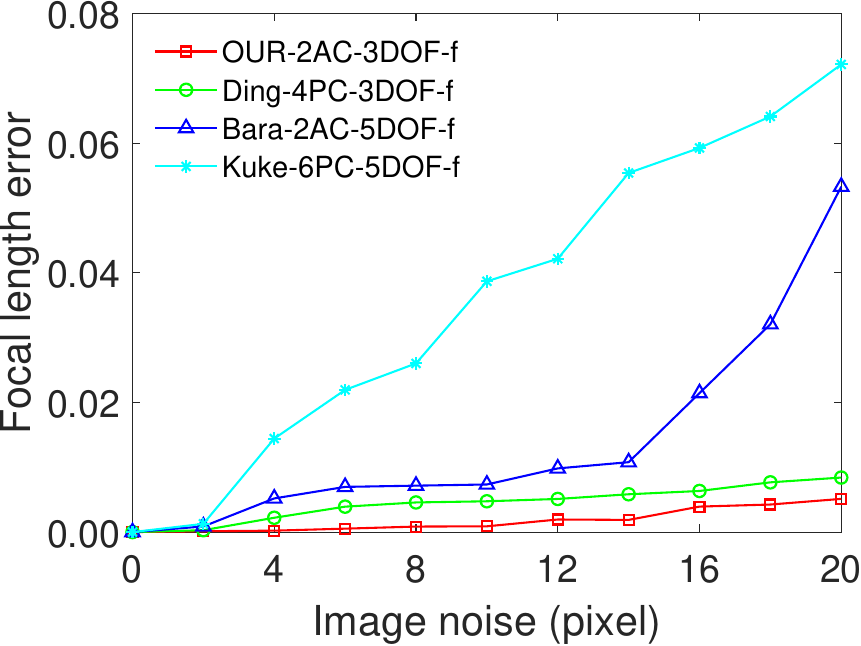}
     \end{minipage}%
      }%
      \subfloat[]{
     \begin{minipage}[t]{0.33\linewidth}
     \centering
     \includegraphics[width=0.9\linewidth]{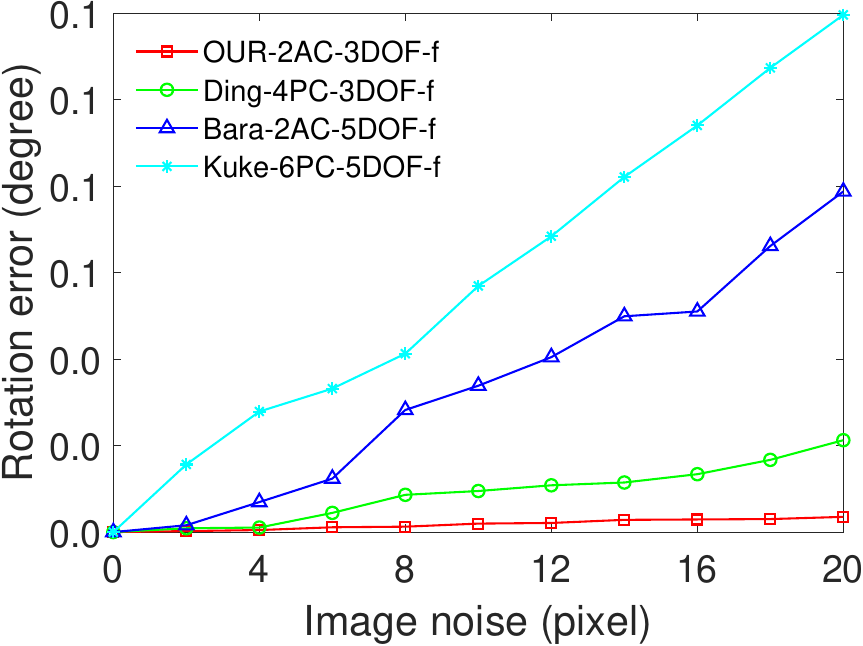}
     \end{minipage}%
      }%
      \subfloat[]{
     \begin{minipage}[t]{0.33\linewidth}
     \centering
     \includegraphics[width=0.9\linewidth]{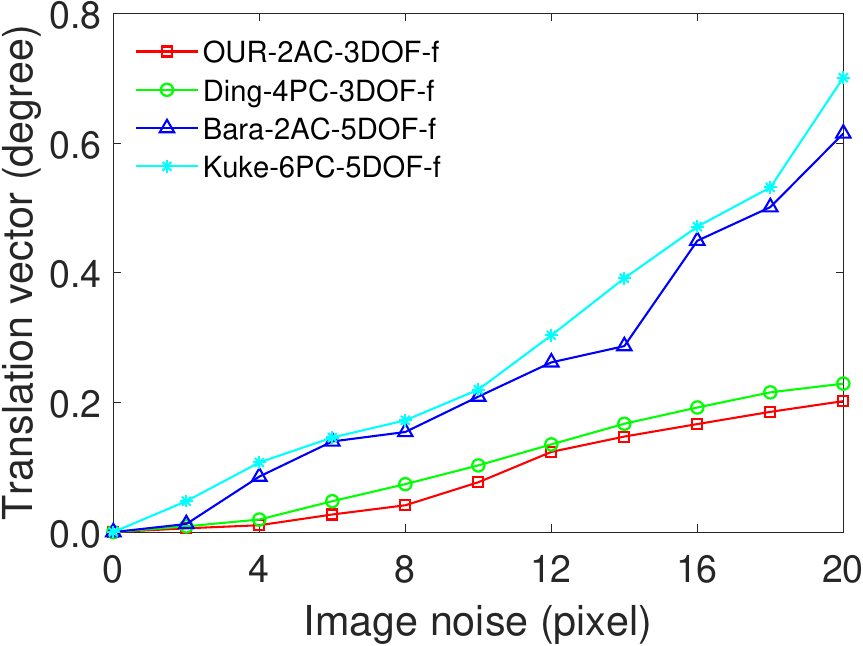}
     \end{minipage}%
      }%
    \centering
    \caption{ \textcolor{black}{Algorithm performance under principal point noise during random motion. (a) Focal length error, (b) Rotation error, (b) Translation error.}}
    \label{principal}
\end{figure}

A sensitivity analysis is conducted to investigate the effect of principal point noise under random motion. Additive noise in the range of 0-20 pixels (with a 2-pixel step) is applied to the principal point, while all other factors are kept noise-free. The experimental results are shown in Fig.~\ref{principal}. The results indicate that our method (shown in red) exhibits notably lower errors in both relative pose and focal length estimation under principal point noise compared to other methods. This performance, consistent across the tested noise range, validates the advantage of our method.

\subsection{Experiments on KITTI dataset}
In order to demonstrate the superiority of the proposed method, we chose the KITTI dataset~\cite{KITTI}. The KITTI dataset is a collection created by the Toyota Technological Institute at Chicago (TTIC) and the Karlsruhe Institute of Technology (KIT) in Germany. It is one of the important datasets in the field of autonomous driving. The data collection platform consists of two color cameras, two grayscale cameras, four optical cameras, a 64-line 3D LiDAR, and an inertial navigation system. The dataset includes a total of 22 sequences, primarily covering five types of scenes: roads, cities, residential areas, campuses, and people. However, only 11 of these sequences (00-10) contain ground truth data, including high-precision GPS values and true motion parameters, while the latter 11 sequences do not provide ground truth for the motion parameters. Therefore, the 11 sequences with ground truth data are selected as validation data for this chapter. These 11 sequences contain a total of 23,190 pairs of images. Feature points and their affine transformations are extracted and matched across image pairs using the ASIFT algorithm~\cite{ASIFT}.

\begin{table}[htbp]
  \centering
  \caption{The error of focal length, rotation matrix, and translation vector on KITTI data.}
  	\scalebox{0.6}{
    \begin{tabular}{rrrrrrrrrrrrr}
    \hline
    \multirow{2}{*}{seq} & \multicolumn{3}{c}{Kuke-6PC-5DOF-f} & \multicolumn{3}{c}{Bara-2AC-5DOF-f} & \multicolumn{3}{c}{Ding-4PC-3DO-f} & \multicolumn{3}{c}{OUR-2AC-3DOF-f}\\
\cline{2-13}  & \multicolumn{1}{c}{${\varepsilon _f}$} & \multicolumn{1}{c}{${\varepsilon _{\bf{R}}}$} & \multicolumn{1}{c}{${\varepsilon _{\bf{t}}}$} & \multicolumn{1}{c}{${\varepsilon _f}$} & \multicolumn{1}{c}{${\varepsilon _{\bf{R}}}$} & \multicolumn{1}{c}{${\varepsilon _{\bf{t}}}$} & \multicolumn{1}{c}{${\varepsilon _f}$} & \multicolumn{1}{c}{${\varepsilon _{\bf{R}}}$} & \multicolumn{1}{c}{${\varepsilon _{\bf{t}}}$} & \multicolumn{1}{c}{${\varepsilon _f}$} & \multicolumn{1}{c}{${\varepsilon _{\bf{R}}}$} & \multicolumn{1}{c}{${\varepsilon _{\bf{t}}}$}\\
    \hline
      00 &	0.208 &	0.935 &	3.642 &	0.170 &	0.827 &	3.213 &	0.101 &	0.167 &	1.756 &	\textbf{0.090} &\textbf{0.124} &\textbf{1.312}\\
      01 &	0.285 &	1.163 &	4.616 &	0.206 &	0.930 &	4.060 &	0.116 &	0.139 &	2.147 &	\textbf{0.084} &\textbf{0.108} &\textbf{1.811}\\
      02 &	0.236 &	1.123 &	3.204 &	0.203 &	1.025 &	2.459 &	0.092 &	0.192 &	1.981 &	\textbf{0.097} &\textbf{0.142} &\textbf{1.714}\\
      03 &	0.244 &	1.144 &	4.058 &	0.216 &	0.926 &	3.751 &	0.090 &	0.166 &	1.923 &	\textbf{0.070} &\textbf{0.104} &\textbf{1.753}\\
      04 &	0.195 &	0.600 &	4.780 &	0.151 &	0.600 &	3.959 &	\textbf{0.068} &\textbf{0.067} &\textbf{2.212} &\textbf{0.068} &0.071 &	2.303\\
      05 &	0.203 &	0.735 &	2.265 &	0.188 &	0.702 &	2.032 &	0.089 &	0.157 &	1.606 &	\textbf{0.077} &\textbf{0.111} &\textbf{1.376}\\
      06 &	0.187 &	0.688 &	3.367 &	0.198 &	0.519 &	2.954 &	0.078 &	0.121 &	1.647 &	\textbf{0.076} &\textbf{0.086} &\textbf{1.272}\\
      07 &	0.195 &	0.655 &	4.121 &	0.167 &	0.510 &	3.655 &	0.072 &	0.168 &	1.948 &	\textbf{0.068} &\textbf{0.127} &\textbf{1.628}\\
      08 &	0.215 &	0.763 &	4.445 &	0.176 &	0.623 &	3.706 &	0.090 &	0.155 &	2.078 &	\textbf{0.073} &\textbf{0.116} &\textbf{1.742}\\
      09 &	0.182 &	1.022 &	2.910 &	0.170 &	0.920 &	2.218 &	0.099 &\textbf{0.199} &	1.764 &	\textbf{0.064} &\textbf{0.199} &\textbf{1.553}\\
      10 &	0.224 &	0.912 &	2.804 &	0.187 &	0.803 &	2.597 &	0.100 &	0.152 &	1.817 &	\textbf{0.070} &\textbf{0.118} &\textbf{1.801}\\
       \hline
     AVG &	\textcolor{black}{0.216} &	\textcolor{black}{0.885} &	\textcolor{black}{3.653} &	\textcolor{black}{0.185} &	\textcolor{black}{0.762} &	\textcolor{black}{3.146} &	\textcolor{black}{0.091} &	\textcolor{black}{0.153} &	\textcolor{black}{1.898} &	\textcolor{black}{\textbf{0.076}} &\textcolor{black}{\textbf{0.119}} &\textcolor{black}{\textbf{1.660}}\\
    \hline
    \end{tabular}}%
  \label{tab2}%
\end{table}%

In Table~\ref{tab2}, the bold black font indicates the smallest error values calculated within the same sequences. The proposed \texttt{OUR-2AC-3DOF-f} method yields focal length errors that are smaller than those of the other three comparison methods (for sequences 00, 01, 02, 04, 05, 06, 07, 08, and 10). The rotational matrix and translation vector errors computed by the \texttt{OUR-2AC-3DOF-f} method are also smaller than those from the other three comparative methods (except for sequence 04). \textcolor{black}{On sequence 04, the focal length error calculated by our method is equal to that calculated by the \texttt{Ding-4PC-3DOF-f} method, while the translation vector error and rotation matrix error are slightly larger. On sequence 09, the rotation matrix error calculated by our method is equal to that calculated by the \texttt{Ding-4PC-3DOF-f} method. The last column of the table shows the mean values, which demonstrates that the proposed method yields the smallest error.}

\subsection{Experiments on smartphone dataset}
Six sets of data were collected using a smartphone. The resolution of the images is 960×1280, with a capture frequency of 15 fps, while the IMU (Inertial Measurement Unit) has a sampling frequency of 105 Hz. The IMU data was aligned with the images through interpolation. The ground truth for the relative rotation matrices and relative translation vectors were obtained using COLMAP software. 

\begin{table}[htbp]
  \centering
  \caption{The error of focal length, rotation matrix, and translation vector on smartphone data.}
  	\scalebox{0.6}{
    \begin{tabular}{rrrrrrrrrrrrr}
    \hline
    \multirow{2}{*}{seq} & \multicolumn{3}{c}{Kuke-6PC-5DOF-f} & \multicolumn{3}{c}{Bara-2AC-5DOF-f} & \multicolumn{3}{c}{Ding-4PC-3DO-f} & \multicolumn{3}{c}{OUR-2AC-3DOF-f}\\
\cline{2-13}  & \multicolumn{1}{c}{${\varepsilon _f}$} & \multicolumn{1}{c}{${\varepsilon _{\bf{R}}}$} & \multicolumn{1}{c}{${\varepsilon _{\bf{t}}}$} & \multicolumn{1}{c}{${\varepsilon _f}$} & \multicolumn{1}{c}{${\varepsilon _{\bf{R}}}$} & \multicolumn{1}{c}{${\varepsilon _{\bf{t}}}$} & \multicolumn{1}{c}{${\varepsilon _f}$} & \multicolumn{1}{c}{${\varepsilon _{\bf{R}}}$} & \multicolumn{1}{c}{${\varepsilon _{\bf{t}}}$} & \multicolumn{1}{c}{${\varepsilon _f}$} & \multicolumn{1}{c}{${\varepsilon _{\bf{R}}}$} & \multicolumn{1}{c}{${\varepsilon _{\bf{t}}}$}\\
    \hline
     01	&0.262	&1.002	&5.951	&0.225	&0.945	&4.942	&0.111	&0.290	&2.707	&\textbf{0.101}	&\textbf{0.206}	&\textbf{2.056}\\
     02	&0.272	&1.123	&7.019	&0.248	&1.117	&5.981	&0.147	&0.513	&3.269	&\textbf{0.130}	&\textbf{0.407}	&\textbf{2.308}\\
     03	&0.265	&1.132	&6.999	&0.231	&1.028	&5.326	&0.146	&0.472	&2.923	&\textbf{0.123}	&\textbf{0.325}	&\textbf{2.210}\\
     04	&0.234	&1.182	&6.243	&0.229	&1.101	&5.032	&0.128	&0.329	&2.587	&\textbf{0.111}	&\textbf{0.292}	&\textbf{2.123}\\
     05	&0.258	&1.364	&6.427	&0.243	&1.263	&5.626	&0.139	&0.438	&3.025	&\textbf{0.129}	&\textbf{0.361}	&\textbf{2.329}\\
     06	&0.276	&1.502	&7.054	&0.246	&1.354	&6.048	&0.149	&0.521	&3.176	&\textbf{0.128}	&\textbf{0.435}	&\textbf{2.418}\\
     \hline
    \textcolor{black}{AVG}	&\textcolor{black}{0.261}	&\textcolor{black}{1.218}	&\textcolor{black}{6.616}	&\textcolor{black}{0.237}	&\textcolor{black}{1.135}	&\textcolor{black}{5.493}	&\textcolor{black}{0.137}	&\textcolor{black}{0.427}	&\textcolor{black}{2.948}	&\textcolor{black}{\textbf{0.120}}	&\textcolor{black}{\textbf{0.338}}	&\textcolor{black}{\textbf{2.241}}\\
    \hline
    \end{tabular}}%
  \label{tab3}%
\end{table}%
Table~\ref{tab3} shows the error results of the focal length, rotation matrix, and translation vector estimation on the smartphone dataset. The errors of the focal length, rotation matrix and translation vector calculated by the \texttt{OUR-2AC-3DOF-f} method are all smaller than those of the comparison method. \textcolor{black}{The overall estimation errors for the rotation matrix and translation vector in our dataset are higher than those in the KITTI dataset—primarily due to the challenging rainy conditions that led to more outlier correspondences and less accurate feature extraction.}

\subsection{Experiments on vehicular mobile platform dataset}

\textcolor{black}{Two sets of data were collected using a vehicular mobile platform. Sensors comprised a LiDAR unit, a radar, an INS, a stereo camera, and an integrated GNSS/INS system. We collected two data sequences, each spanning roughly 2 kilometers. We develop an alternative version of our solver using the Gröbner basis method instead of the polynomial eigenvalue approach, named \texttt{OUR-2AC-Gröbner}. The results are shown in Table~\ref{tab4}}. \textcolor{black}{The proposed method demonstrates superior performance by achieving the smallest relative pose and focal length errors on our challenging rainy-weather dataset.}

\begin{table}[tp]
  \centering
  \caption{The error of focal length, rotation matrix, and translation vector on vehicular mobile platform data.}
  	\scalebox{0.53}{
    \begin{tabular}{rrrrrrrrrrrrrrrr}
    \hline
    \multirow{2}{*}{seq} & \multicolumn{3}{c}{Kuke-6PC-5DOF-f} & \multicolumn{3}{c}{Bara-2AC-5DOF-f} & \multicolumn{3}{c}{Ding-4PC-3DO-f} & \multicolumn{3}{c}{OUR-2AC-Gröbner} & \multicolumn{3}{c}{OUR-2AC-3DOF-f}\\
\cline{2-16}  & \multicolumn{1}{c}{${\varepsilon _f}$} & \multicolumn{1}{c}{${\varepsilon _{\bf{R}}}$} & \multicolumn{1}{c}{${\varepsilon _{\bf{t}}}$} & \multicolumn{1}{c}{${\varepsilon _f}$} & \multicolumn{1}{c}{${\varepsilon _{\bf{R}}}$} & \multicolumn{1}{c}{${\varepsilon _{\bf{t}}}$} & \multicolumn{1}{c}{${\varepsilon _f}$} & \multicolumn{1}{c}{${\varepsilon _{\bf{R}}}$} & \multicolumn{1}{c}{${\varepsilon _{\bf{t}}}$} & \multicolumn{1}{c}{${\varepsilon _f}$} & \multicolumn{1}{c}{${\varepsilon _{\bf{R}}}$} & \multicolumn{1}{c}{${\varepsilon _{\bf{t}}}$}& \multicolumn{1}{c}{${\varepsilon _f}$} & \multicolumn{1}{c}{${\varepsilon _{\bf{R}}}$} & \multicolumn{1}{c}{${\varepsilon _{\bf{t}}}$}\\
    \hline
     01	&0.292	&1.523	&7.825	&0.325	&1.335	&6.442	&0.153	&0.579	&3.707 &0.168&0.556&3.112   &\textbf{0.152}	&\textbf{0.524}	&\textbf{2.862}\\
     02	&0.282	&1.511	&7.228	&0.298	&1.253	&6.021	&0.149	&0.535	&3.689 &0.149 &0.527 &3.008	&\textbf{0.144}	&\textbf{0.486}	&\textbf{2.788}\\
     \hline
    AVG	& 0.287	&  1.517 & 7.527 & 0.312	&1.294	&6.232	&0.151	&0.557	& 3.697&0.159&0.542&3.600	&\textbf{0.148}	&\textbf{0.505}	&\textbf{2.825}\\
    \hline
    \end{tabular}}%
  \label{tab4}%
\end{table}%

\begin{table}[tp]
  \centering
  \caption{\textcolor{black}{Runtime of RANSAC averaged over vehicular mobile platform data (unit: ${s}$)}}
  	\scalebox{0.66}{
    \begin{tabular}{cccccc}
    \hline
    method & Kuke-6PC-5DOF-f & Bara-2AC-5DOF-f & Ding-4PC-3DO-f& OUR-2AC-Gröbner & OUR-2AC-3DOF-f\\
    \hline
     time & 2.55 & 1.81 & 0.10 & 0.80& 0.05\\
    \hline 
    \end{tabular}}%
  \label{tab5}%
\end{table}%

Table~\ref{tab5} presents the average RANSAC runtimes of all methods on the vehicular mobile 
platform data. The results demonstrate that the \texttt{OUR-2AC-3DOF-f} method consumes the shortest time.

\section{Conclusion}
In this paper, we propose a new solver to estimate the relative pose and focal length between two views for the semi-calibrated camera with a known vertical direction. This assumption is very common in practical applications. We establish the constraint equation from affine correspondences. A polynomial eigenvalue solver is employed to estimate the focal length, rotation matrix, and translation vector. The numerical stability of our solver is better than that of the comparison method in synthetic data. Besides, the performance of estimating focal length, rotation matrix, and translation vector calculated by our method is better than that of the comparison method in synthetic data and real data. Our method can be applied to autonomous driving, unmanned aerial vehicles, sweeping robots, and other fields.

\bibliographystyle{IEEEtran}
\bibliography{references}

@inproceedings{SMF1,
  title={Structure-from-Motion Revisited},
  author={ Sch{\"o}nberger,Johannes L. and Frahm, Jan-Michael.},
  booktitle={Proc. IEEE Conf. Comput. Vis. Pattern Recognit. (CVPR)},
  pages={4104-4113},
  year={2016},
}

@article{VO,
  title={City-Scale Localization for Cameras with Known Vertical Direction},
  author={ Svarm, L.  and  Enqvist, O.  and  Kahl, F.  and  Oskarsson, M. },
  journal={IEEE Trans. Pattern Anal. Mach. Intell.},
  volume={39},
  number={7},
  pages={1455-1461},
  year={2017},
}

@ARTICLE{VO_tim,
  author={Chiodini, Sebastiano and Giubilato, Riccardo and Pertile, Marco and Debei, Stefano},
  journal={IEEE Trans. Instrum. Meas.}, 
  title={Retrieving Scale on Monocular Visual Odometry Using Low-Resolution Range Sensors}, 
  year={2020},
  volume={69},
  number={8},
  pages={5875-5889},
  }

@INPROCEEDINGS{SFM2,
  author={Raposo, Carolina and Barreto, João P.},
  booktitle={Proc. IEEE Conf. Comput. Vis. Pattern Recognit. (CVPR)}, 
  title={Theory and Practice of Structure-From-Motion Using Affine Correspondences}, 
  year={2016},
  volume={},
  number={},
  pages={5470-5478}}

@ARTICLE{SLAM2_RAL,
  author={Mur-Artal, Raúl and Tardós, Juan D.},
  journal={IEEE Trans. Robot}, 
  title={ORB-SLAM2: An Open-Source SLAM System for Monocular, Stereo, and RGB-D Cameras}, 
  year={2017},
  volume={33},
  number={5},
  pages={1255-1262}}

@ARTICLE{SLAM3_RAL,
  title={Orb-slam3: An accurate open-source library for visual, visual--inertial, and multi--map slam},
  author={Campos, Carlos and Elvira, Richard and Rodr{\'\i}guez, Juan J G{\'o}mez and Montiel, Jos{\'e} M M and Tard{\'o}s, Juan D},
  journal={IEEE Trans. Robot},
  volume={37},
  number={6},
  pages={1874--1890},
  year={2021},
}

@INPROCEEDINGS{guan1,
  author={Guan, Banglei and Zhao, Ji and Li, Zhang and Sun, Fang and Fraundorfer, Friedrich},
  booktitle={Proc. IEEE Conf. Comput. Vis. Pattern Recognit. (CVPR)}, 
  title={Minimal Solutions for Relative Pose With a Single Affine Correspondence}, 
  year={2020},
 pages={1926-1935},
 }

@article{ASIFT,
  title={ASIFT: A new framework for fully affine invariant image comparison},
  author={ Morel, J. M.  and  Yu, G. },
  journal={SIAM J. Imag. Sci.},
  volume={2},
  number={2},
  pages={438-469},
  year={2009},
}

@INPROCEEDINGS{2ac,
  author={Barath, Daniel and Toth, Tekla and Hajder, Levente},
  booktitle={Proc. IEEE Conf. Comput. Vis. Pattern Recognit. (CVPR)}, 
  title={A Minimal Solution for Two-View Focal-Length Estimation Using Two Affine Correspondences}, 
  year={2017},
  volume={},
  number={},
  pages={2557-2565},
 }

@article{Random,
  title={Random Sample Consensus: A Paradigm for Model Fitting with Applications to Image Analysis and Automated Cartography-ScienceDirect},
  author={ Fischler, M. A.  and  Bolles, R. C. },
  journal={Commun, ACM},
  volume={24},
  number={6},
  pages={381-395},
  year={1981},
}

@ARTICLE{Polynomial,
  author={Kukelova, Zuzana and Bujnak, Martin and Pajdla, Tomas},
  journal={IEEE Trans. Pattern Anal. Mach. Intell.}, 
  title={Polynomial Eigenvalue Solutions to Minimal Problems in Computer Vision}, 
  year={2012},
  volume={34},
  number={7},
  pages={1381-1393}}

@ARTICLE{5PT1,
  author={Nister, D.},
  journal={IEEE Trans. Pattern Anal. Mach. Intell.}, 
  title={An efficient solution to the five-point relative pose problem}, 
  year={2004},
  volume={26},
  number={6},
  pages={756-770}}

@INPROCEEDINGS{IMU1,
  title={A minimal case solution to the calibrated relative pose problem for the case of two known orientation angles},
  author={Fraundorfer,Friedrich and Tanskanen,  Petri and Pollefeys, Marc},
  booktitle={Proc. European Conference on Computer Vision. (ECCV)},
  pages={269--282},
  year={2010},
}

@inproceedings{IMU2,
 author={Kneip, Laurent and Chli, Margarita and Siegwart, Roland},
  booktitle={Proc. British Machine Vision Conference. (BMVC)}, 
  title={Robust real-time visual odometry with a single camera and an IMU}, 
  year={2011},
  volume={ },
  number={ },
  pages={1-12}
}

@inproceedings{IMU3,
 author={Sweeney, Chris and Flynn, John and Turk, Matthew},
  booktitle={Proc. International Conference on 3D Vision. (3DV)}, 
  title={Solving for relative pose with a partially known rotation is a quadratic eigenvalue problem}, 
  year={2014},
 pages={483--490}
}

@INPROCEEDINGS{IMU4,
  author={Guan, Banglei and Vasseur, Pascal and Demonceaux, Cedric and Fraundorfer, Friedrich},
  booktitle={Proc. Int. Conf. Robot. Autom. (ICRA)}, 
  title={Visual Odometry Using a Homography Formulation with Decoupled Rotation and Translation Estimation Using Minimal Solutions}, 
  year={2018},
  pages={2320-2327}}

@article{KITTI,
  title={Vision meets robotics: The {KITTI} dataset},
  author={Geiger, Andreas and Lenz, Philip and Stiller},
  journal={The International Journal of Robotics Research},
  volume={32},
  number={11},
  pages={1231--1237},
  year={2013},
  publisher={Sage Publications Sage UK: London, England}
   }

@INPROCEEDINGS{AC2,
  author={Barath, Daniel and Toth, Tekla and Hajder, Levente},
  booktitle={Proc. IEEE Conf. Comput. Vis. Pattern Recognit. (CVPR)}, 
  title={A Minimal Solution for Two-View Focal-Length Estimation Using Two Affine Correspondences}, 
  year={2017},
  volume={},
  number={},
  pages={2557-2565},
}

@INPROCEEDINGS{barath_planar,
  author={Hajder, Levente and Barath, Daniel},
  booktitle={Proc. Int. Conf. Robot. Autom. (ICRA)}, 
  title={Relative planar motion for vehicle-mounted cameras from a single affine correspondence}, 
  year={2020},
  volume={},
  number={},
  pages={8651-8657},
}

@inproceedings{lihongdong,
  title={A simple solution to the six-point two-view focal-length problem},
  author={Li, Hongdong},
  booktitle={Proc. European Conference on Computer Vision. (ECCV)},
  pages={200--213},
  year={2006},
}

@article{stewenius,
  title={A minimal solution for relative pose with unknown focal length},
  author={Stew{\'e}nius, Henrik and Nist{\'e}r, David and Kahl, Fredrik and Schaffalitzky, Frederik},
  journal={Int. J. Comput. Vis.},
  volume={26},
  number={7},
  pages={871--877},
  year={2008},
  publisher={Elsevier}
}

@book{book,
  title={Multiple view geometry in computer vision},
  author={Hartley, Richard and Zisserman, Andrew},
  pages={187--198},
  year={2003},
  publisher={Cambridge university press}
}

@inproceedings{stewenius_10,
  title={Critical motions in euclidean structure from motion},
  author={Kahl, Fredrik and Triggs, Bill},
  booktitle={Proc. IEEE Conf. Comput. Vis. Pattern Recognit. (CVPR)},
  volume={2},
  pages={366--372},
  year={1999},
}

@inproceedings{stewenius_21,
  title={On focal length calibration from two views},
  author={Sturm, Peter},
  booktitle={Proc. IEEE Conf. Comput. Vis. Pattern Recognit. (CVPR)},
  volume={2},
  pages={1--8},
  year={2001},
}

@inproceedings{problem_6pt,
  title={The six point algorithm revisited},
  author={Torii, Akihiko and Kukelova, Zuzana and Bujnak, Martin and Pajdla, Tomas},
  booktitle={Proc. Asian Conference on Computer Vision. (ACCV)},
  pages={184--193},
  year={2010},
}

@inproceedings{ding_f,
  title={Minimal solutions to relative pose estimation from two views sharing a common direction with unknown focal length},
  author={Ding, Yaqing and Yang, Jian and Ponce, Jean and Kong, Hui},
  booktitle={Proc. IEEE Conf. Comput. Vis. Pattern Recognit. (CVPR)},
  pages={7045--7053},
  year={2020}
}

@inproceedings{ding_H,
  title={An efficient solution to the homography-based relative pose problem with a common reference direction},
  author={Ding, Yaqing and Yang, Jian and Ponce, Jean and Kong, Hui},
  booktitle={Proc. IEEE Int. Conf. Comput. Vis. (ICCV)},
  pages={1655--1664},
  year={2019},
}

@inproceedings{barath_homography,
  title={Homography from two orientation-and scale-covariant features},
  author={Barath, Daniel and Kukelova, Zuzana},
  booktitle={Proc. Int. Conf. Comput. Vis.},
  pages={1091--1099},
  year={2019}
}

@article{bentolila2014conic,
  title={Conic epipolar constraints from affine correspondences},
  author={Bentolila, Jacob and Francos, Joseph M},
  journal={Comput. Vis. Image Understand.},
  volume={122},
  pages={105--114},
  year={2014}
}

@article{barath2018Essential,
  title={Efficient recovery of essential matrix from two affine correspondences},
  author={Barath, Daniel and Hajder, Levente},
  journal={IEEE Trans. Image Process.},
  volume={27},
  number={11},
  pages={5328--5337},
  year={2018},
  publisher={IEEE}
}

@inproceedings{eichhardt2018affine,
  title={Affine correspondences between central cameras for rapid relative pose estimation},
  author={Eichhardt, Iv{\'a}n and Chetverikov, Dmitry},
  booktitle={Proc. European Conference on Computer Vision. (ECCV)},
  pages={482--497},
  year={2018}
}

@inproceedings{kukelova2008Gröbner,
  title={Automatic generator of minimal problem solvers},
  author={Kukelova, Zuzana and Bujnak, Martin and Pajdla, Tomas},
  booktitle={Proc. European Conference on Computer Vision. (ECCV)},
  pages={302--315},
  year={2008},
}

@book{eigenvalue,
  title={Templates for the solution of algebraic eigenvalue problems: a practical guide},
  author={Bai, Zhaojun and Demmel, James and Dongarra, Jack and Ruhe, Axel and van der Vorst, Henk},
  year={2000},
  publisher={Society for Industrial and Applied Mathematics}
}

@inproceedings{kukelova20093d,
  title={3D reconstruction from image collections with a single known focal length},
  author={Bujnak, Martin and Kukelova, Zuzana and Pajdla, Tomas},
  booktitle={Proc. IEEE Int. Conf. Comput. Vis. (ICCV)},
  pages={1803--1810},
  year={2009},
  organization={IEEE}
}

@article{saurer2016homography,
  title={Homography based egomotion estimation with a common direction},
  author={Saurer, Olivier and Vasseur, Pascal and Boutteau, R{\'e}mi and Demonceaux, C{\'e}dric and Pollefeys, Marc and Fraundorfer, Friedrich},
  journal={IEEE Trans. Pattern Anal. Mach. Intell.},
  volume={39},
  number={2},
  pages={327--341},
  year={2016},
  publisher={IEEE}
}

@inproceedings{fitzgibbon2001simultaneous,
  title={Simultaneous linear estimation of multiple view geometry and lens distortion},
  author={Fitzgibbon, Andrew W},
  booktitle={Proc. IEEE Conf. Comput. Vis. Pattern Recognit. (CVPR)},
  volume={1},
  pages={I--I},
  year={2001},
}

@inproceedings{larsson2018beyond,
  title={Beyond grobner bases: Basis selection for minimal solvers},
  author={Larsson, Viktor and Oskarsson, Magnus and Astrom, Kalle and Wallis, Alge and Kukelova, Zuzana and Pajdla, Tomas},
  booktitle={Proc. IEEE Conf. Comput. Vis. Pattern Recognit. (CVPR)},
  pages={3945--3954},
  year={2018}
}

@ARTICLE{vO2_tim,
  author={Ban, Xicheng and Wang, Hongjian and Chen, Tao and Wang, Ying and Xiao, Yao},
  journal={IEEE Trans. Instrum. Meas.}, 
  title={Monocular Visual Odometry Based on Depth and Optical Flow Using Deep Learning}, 
  year={2021},
  volume={70},
  number={},
  pages={1-19},
 }

@article{SFM_TMM,
  title={Improved Image-Based Localization Using SFM and Modified Coordinate System Transfer},
  author={Salarian,Mahdi and Ilievn, Nick and Cetin, Ahmet Enis and Ansari, Rashid},
  journal={IEEE Transactions on Multimedia},
  volume={20},
  number={12},
  pages={3298-3310},
  year={2018},
}

@inproceedings{sun2025learning,
  title={Learning affine correspondences by integrating geometric constraints},
  author={Sun, Pengju and Guan, Banglei and Yu, Zhenbao and Shang, Yang and Yu, Qifeng and Barath, Daniel},
  booktitle={Proc. IEEE Conf. Comput. Vis. Pattern Recognit. (CVPR)},
  pages={27038--27048},
  year={2025}
}

@article{Dosovitskiy2015FlowNetLO,
  title={FlowNet: Learning Optical Flow with Convolutional Networks},
  author={Alexey Dosovitskiy and Philipp Fischer and Eddy Ilg and Philip H{\"a}usser and Caner Hazirbas and Vladimir Golkov and Patrick van der Smagt and Daniel Cremers and Thomas Brox},
  journal={Proc. IEEE Int. Conf. Comput. Vis.},
   pages={2758-2766},
  year={2015}
 }

@inproceedings{mayer2016large,
  title={A large dataset to train convolutional networks for disparity, optical flow, and scene flow estimation},
  author={Mayer, Nikolaus and Ilg, Eddy and Hausser, Philip and Fischer, Philipp and Cremers, Daniel and Dosovitskiy, Alexey and Brox, Thomas},
  booktitle={Proc. IEEE Conf. Comput. Vis. Pattern Recognit. (CVPR)},
  pages={4040--4048},
  year={2016}
}

@article{vijayanarasimhan1704sfm,
  title={Sfm-net: Learning of structure and motion from video. arXiv 2017},
  author={Vijayanarasimhan, S and Ricco, S and Schmid, C and Sukthankar, R and Fragkiadaki, K},
  journal={arXiv preprint arXiv:1704.07804}
}

@inproceedings{wang2021deep,
  title={Deep two-view structure-from-motion revisited},
  author={Wang, Jianyuan and Zhong, Yiran and Dai, Yuchao and Birchfield, Stan and Zhang, Kaihao and Smolyanskiy, Nikolai and Li, Hongdong},
  booktitle={Proc. IEEE Conf. Comput. Vis. Pattern Recognit. (CVPR)},
  pages={8953--8962},
  year={2021}
}

@inproceedings{zhuang2021fusing,
  title={Fusing the old with the new: Learning relative camera pose with geometry-guided uncertainty},
  author={Zhuang, Bingbing and Chandraker, Manmohan},
  booktitle={Proc. IEEE Conf. Comput. Vis. Pattern Recognit. (CVPR)},
  pages={32--42},
  year={2021}
}

@inproceedings{parameshwara2022diffposenet,
  title={Diffposenet: Direct differentiable camera pose estimation},
  author={Parameshwara, Chethan M and Hari, Gokul and Ferm{\"u}ller, Cornelia and Sanket, Nitin J and Aloimonos, Yiannis},
  booktitle={Proc. IEEE Conf. Comput. Vis. Pattern Recognit. (CVPR)},
  pages={6845--6854},
  year={2022}
}

@inproceedings{2013Unified,
  title={Unified temporal and spatial calibration for multi-sensor systems},
  author={ Furgale, Paul  and  Rehder, Joern  and  Siegwart, Roland },
  booktitle={IEEE International Conference on Intelligent Robots Systems},
  pages={1280--1286},
  year={2013},
}

@article{Yang2017Monocular,
  title={Monocular Visual-Inertial State Estimation With Online Initialization and Camera-IMU Extrinsic Calibration},
  author={Yang and Zhenfei and Shen and Shaojie},
  journal={IEEE transactions on automation science and engineering: a publication of the IEEE Robotics and Automation Society},
  volume={14},
  number={1},
  pages={39-51},
  year={2017},
}
\end{document}